\documentclass[]{bytedance}
\usepackage[toc,page,header]{appendix}

\usepackage{minitoc}
\usepackage{amsfonts}
\usepackage{amssymb}
\usepackage{tabularx}
\usepackage{listings}
\usepackage{xcolor}
\usepackage{cancel}

\usepackage{tabulary,multirow,xspace}
\usepackage{fixmath,mathtools,nicefrac,mmstyle}
\usepackage{subcaption}
\usepackage{caption}
\usepackage{wrapfig} % for wrapfigure
\usepackage[misc]{ifsym} % to use \Letter
\usepackage{colortbl}

\usepackage{wrapfig}
\usepackage{multicol}
\usepackage[most]{tcolorbox}
\usepackage{pifont}

\definecolor{codegreen}{rgb}{0,0.6,0}
\definecolor{codegray}{rgb}{0.5,0.5,0.5}
\definecolor{codepurple}{rgb}{0.58,0,0.82}
\definecolor{backcolour}{rgb}{0.95,0.95,0.92}
\definecolor{boxblue}{RGB}{57,89,163}
\definecolor{boxbluebg}{RGB}{230,237,250} 

\lstdefinestyle{mystyle}{
    backgroundcolor=\color{backcolour},   
    commentstyle=\color{codegreen},
    keywordstyle=\color{magenta},
    numberstyle=\tiny\color{codegray},
    stringstyle=\color{codepurple},
    basicstyle=\ttfamily\footnotesize,
    breakatwhitespace=false,         
    breaklines=true,                 
    captionpos=b,                    
    keepspaces=true,                 
    numbers=none,                    
    numbersep=5pt,                  
    showspaces=false,                
    showstringspaces=false,
    showtabs=false,                  
    tabsize=2
}
\definecolor{mygray1}{gray}{.95}
\definecolor{mygray2}{gray}{.9}
\definecolor{mygray3}{gray}{.95}
\usepackage{pifont}% http://ctan.org/pkg/pifont

\newlength\savewidth
\newcolumntype{x}[1]{>{\centering\arraybackslash}p{#1pt}}

\newcommand{\app}{\raise.17ex\hbox{$\scriptstyle\sim$}}

\usepackage{xcolor}
\usepackage{graphicx}
\usepackage{amssymb}
\usepackage{pifont}
\usepackage{floatrow}
\usepackage{amsmath} 
\usepackage{float}
\usepackage{wrapfig}
\usepackage{multirow}
\usepackage{tcolorbox}
\tcbuselibrary{breakable, skins, raster}
\usepackage{listings}
\usepackage{listings}

\usepackage{algorithm}
\usepackage{algorithmic}
\definecolor{myblue}{RGB}{210, 225, 255}
\definecolor{mytextblue}{RGB}{51, 161, 201}
\definecolor{mypurple}{RGB}{218, 112, 214}

\definecolor{commentgreen}{rgb}{0.1, 0.4, 0.1}
\definecolor{keywordblue}{rgb}{0.1, 0.1, 0.7}
\definecolor{stringred}{rgb}{0.7, 0.1, 0.1}

\lstdefinestyle{mystyle}{
    commentstyle=\color{commentgreen},
    keywordstyle=\color{keywordblue},   
    stringstyle=\color{stringred},
    basicstyle=\ttfamily\scriptsize, 
    breaklines=true,
    keepspaces=true,
    showstringspaces=false,
    frame=none,                     
    language=Python, 
}

\newcommand{\name}{USO}
\title{\name{}: Unified Style and Subject-Driven Generation via Disentangled and Reward Learning}

\author{
\centerline{
Shaojin Wu \quad 
Mengqi Huang $^*$ \quad  
Yufeng Cheng \quad 
Wenxu Wu \quad
} 
\centerline{
Jiahe Tian \quad
Yiming Luo \quad
Fei Ding $^{\dagger}$ \quad
Qian He \quad
}
}

\affiliation[]{UXO Team, Intelligent Creation Lab, ByteDance}
\footnotetext{* Corresponding author $\dagger$ Project lead.}
\abstract{
       Existing literature typically treats style-driven and subject-driven generation as two disjoint tasks: the former prioritizes stylistic similarity, whereas the latter insists on subject consistency, resulting in an apparent antagonism. We argue that both objectives can be unified under a single framework because they ultimately concern the disentanglement and re-composition of \textit{content} and \textit{style}, a long-standing theme in style-driven research. To this end, we present \textbf{USO}, a \textbf{U}nified \textbf{S}tyle-\textbf{S}ubject \textbf{O}ptimized customization model. First, we construct a large-scale triplet dataset consisting of content images, style images, and their corresponding stylized content images. Second, we introduce a disentangled learning scheme that simultaneously aligns style features and disentangles content from style through two complementary objectives, style-alignment training and content–style disentanglement training. Third, we incorporate a style reward-learning paradigm denoted as SRL to further enhance the model’s performance. Finally, we release USO-Bench, the first benchmark that jointly evaluates style similarity and subject fidelity across multiple metrics. Extensive experiments demonstrate that USO achieves state-of-the-art performance among open-source models along both dimensions of subject consistency and style similarity. Code and model: \url{https://github.com/bytedance/USO}
}

\date{\today}

\checkdata[Project Page]{\url{https://bytedance.github.io/USO/}}

\correspondence{Shaojin Wu at \email{wushaojin@bytedance.com}}

\begin{document}
\maketitle

\section{Introduction}
\label{sec:intro}

The significant advancements in image generation over the past years have greatly improved generative controllability, fundamentally changing how humans create images, \textit{i.e.}, whether through abstract textual descriptions, specific visual reference images, or both. Research on leveraging both textual and visual conditions has attracted increasing interest, giving rise to numerous real-world tasks such as style-driven generation and subject-driven generation. While textual conditions are typically explicit, \textit{\textbf{visual conditions are inherently noisy}}, as images intrinsically embody a rich spectrum of features (\textit{e.g.}, style, appearance), of which only a specific one is relevant to a specific task. \begin{figure}[H]
\centering  
\vspace*{-0.11\textwidth}
\hspace*{-0.09\textwidth}
\includegraphics[width=1.165\textwidth]{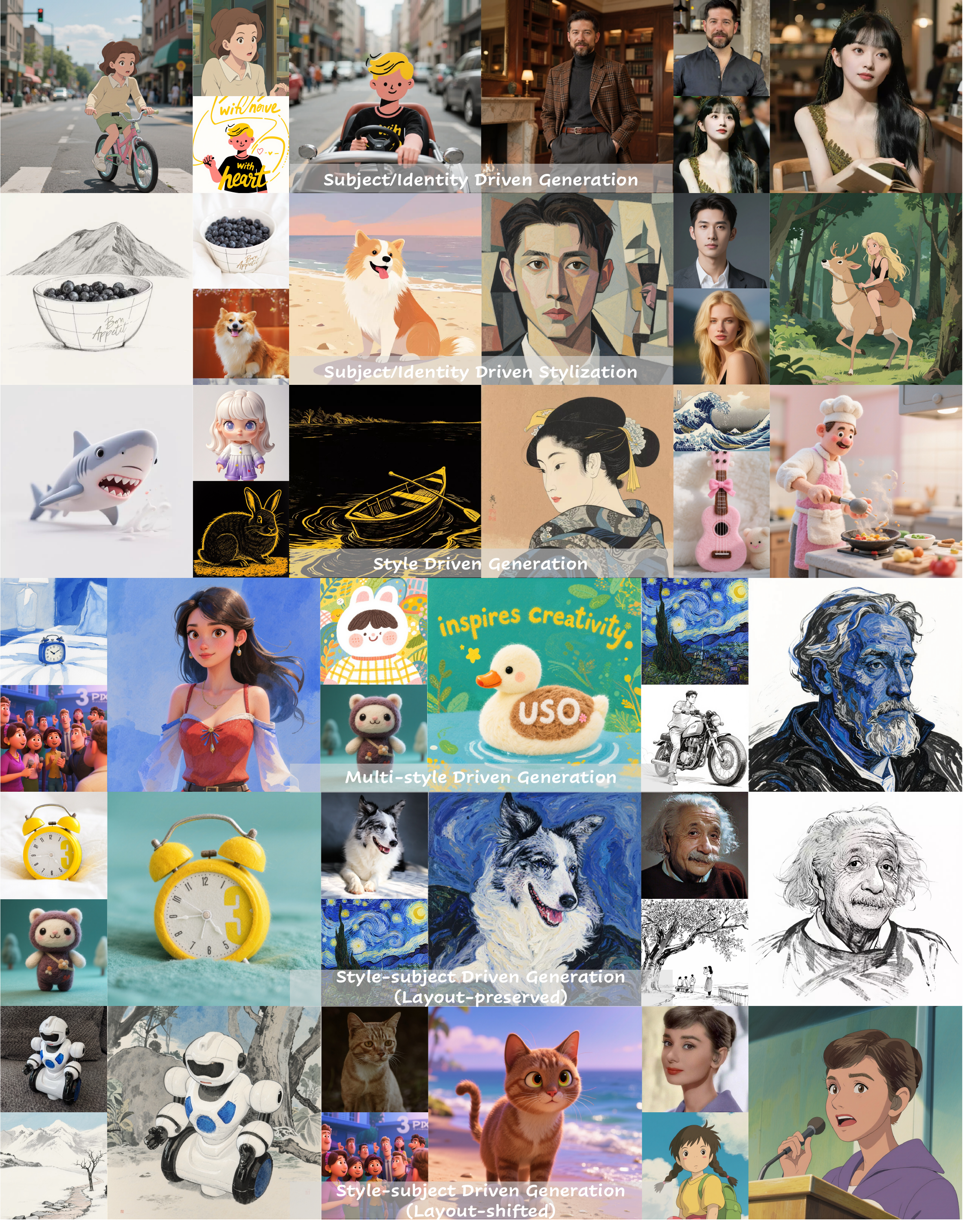}
    \caption{Showcase of the versatile abilities of the \textbf{USO} model. Prompts are in ~\Cref{tab:teaser}.}
    \label{teaser}
\end{figure}For instance, style-driven generation requires only the style feature from the reference images, whereas other features constitute noise. Therefore, a fundamental and long-standing challenge in these tasks is to accurately \textit{\textbf{include all required features from the reference image while simultaneously excluding other noisy ones}}, \textit{e.g.}, including only the style in style-driven generation or only the subject’s appearance in subject-driven generation.

\begin{figure}[t]
\centering
\includegraphics[scale=0.48]{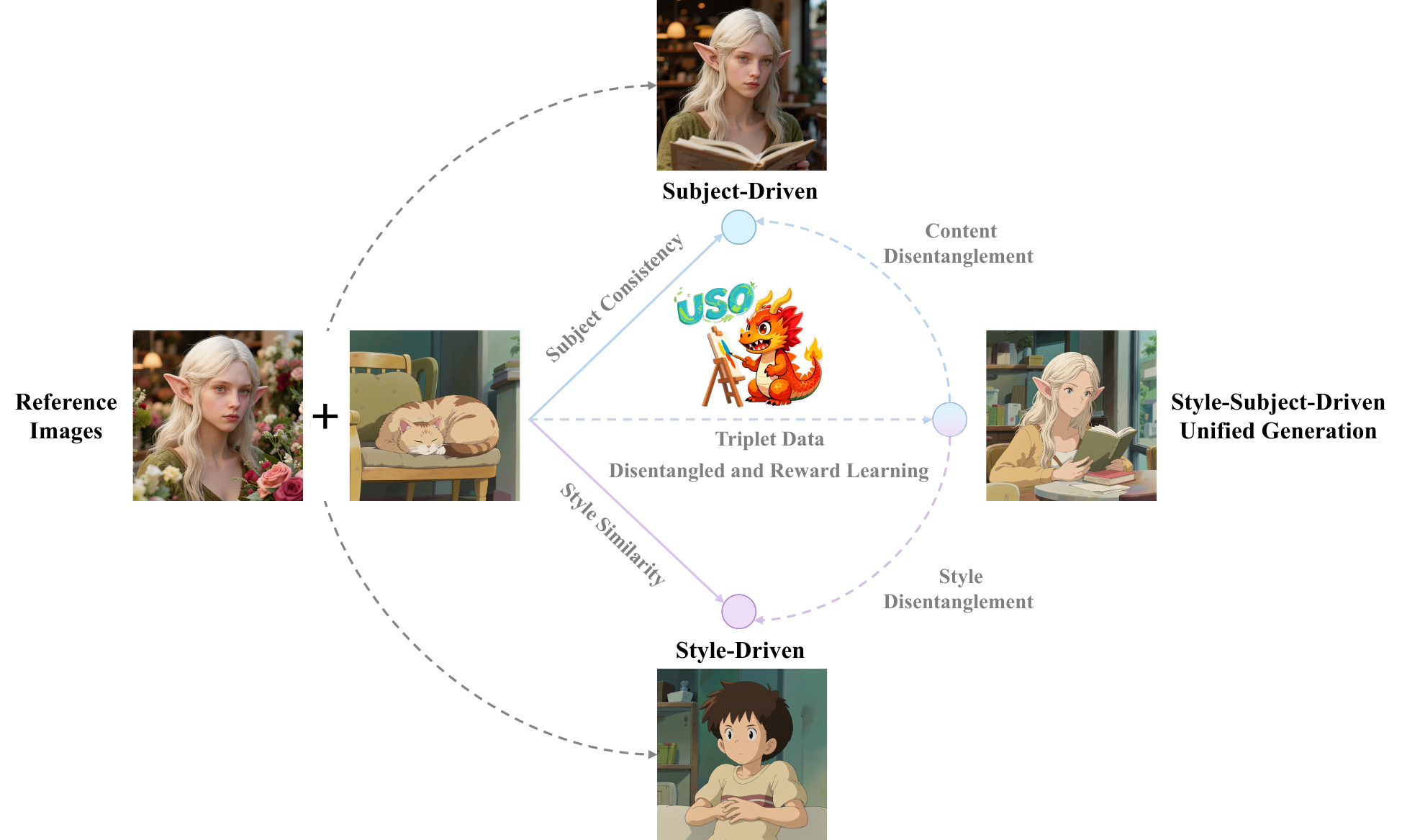}
    \caption{Illustration of our motivation. By jointly disentangling content and style across tasks, we unify style-driven and subject-driven generation within a single framework.}
    \label{fig1}
\end{figure}
Extensive efforts in the literature have been dedicated to disentangling different features in visual conditions (\textit{i.e.}, reference images). On the one hand, in the realm of style-driven generation, DEADiff~\cite{qi2024deadiff} employs QFormer to selectively query only the style features from reference images. CSGO~\cite{xing2024csgo} constructs content-style-stylized triplets to facilitate style-content decoupling during training. StyleStudio~\cite{lei2025stylestudio} introduces style-based classifier-free guidance (SCFG) to enable selective control over stylistic elements and to mitigate the influence of irrelevant features. On the other hand, subject-driven generation methods primarily focus on disentangling subject appearance features or constructing more effectively disentangled paired data. For example, RealCustom~\cite{huang2024realcustom, mao2024realcustom++} proposes a dual-inference framework that selectively incorporates subject-relevant features into subject-specific regions. UNO~\cite{wu2025less} leverages the in-context capabilities of DiT to progressively improve both the quality of paired data and the model itself. To conclude, existing methods primarily focus on \textit{\textbf{task-specific disentanglement}} by designing tailored datasets or model architectures for each individual task, thereby performing \textit{\textbf{disentanglement in an isolated, single-task context.}}

In this study, we argue that a more comprehensive and precise disentanglement approach should fully account for the coupling and complementarity between different generation tasks. Each task should not only learn which features to include, but, more importantly, also learn which features to exclude, \textit{i.e.}, features that are often required by other tasks. \textit{\textbf{Therefore, learning to include certain features in one task inherently informs and enhances the process of learning to exclude those same features in a complementary task, and vice versa.}} For example, style-driven generation aims to incorporate stylistic features while excluding subject appearance features, whereas subject-driven generation does the exact opposite. The ability to learn and include subject appearance features in subject-driven generation can, in turn, help style-driven generation more effectively learn to exclude these features, thereby improving disentanglement for both tasks. In conclusion, we believe that jointly modeling complementary tasks enables a mutually reinforcing disentanglement process, leading to a more precise separation of relevant and irrelevant features for each task.

Based on the above motivation, we propose a novel \textit{\textbf{cross-task co-disentanglement}} paradigm to unify subject-driven and style-driven generation, and, more importantly, to mutually enhance the performance of both tasks, as illustrated in ~\Cref{fig1}. Specifically, this co-disentanglement paradigm is implemented through a \textit{subject-for-style} data curation framework and a \textit{style-for-subject} model training framework. The \textit{subject-for-style} framework first utilizes a state-of-the-art subject model to generate high-quality style data, while the \textit{style-for-subject} framework subsequently trains a more effective subject model under the guidance of style rewards and disentangled training.
Technically, on the one hand, for the \textit{subject-for-style} data curation framework, we build upon a state-of-the-art subject-driven model~\cite{wu2025less} and further develop both a stylization expert and a de-stylization expert to curate stylized and non-stylized images. This process ultimately constructs triplet data pairs in the form of $<$style reference, de-stylized subject reference, stylized subject result$>$ for subsequent model training. On the other hand, for the \textit{style-for-subject} model training framework, we propose a \textbf{U}nified \textbf{S}tyle-\textbf{S}ubject \textbf{O}ptimized (\textbf{USO}) customization model, which introduces progressive style alignment and style-subject disentanglement training, both supervised by a style reward. 

Our contributions are summarized as follows:

\textbf{Concepts: } We point out that existing style-driven and subject-driven methods focus solely on isolated disentanglement within each task, neglecting their potential complementarity and thus leading to suboptimal disentanglement. For the first time, we propose a novel cross-task co-disentanglement paradigm that unifies style-driven and subject-driven tasks, enabling mutual enhancement and achieving significant performance improvements for both.

\textbf{Technique: } We present a novel cross-task triplet curation framework that bridges style-driven and subject-driven generation. Building on this, we introduce USO, a unified customization architecture that incorporates progressive style-alignment training, content–style disentanglement training, and a style reward learning paradigm to further promote cross-task disentanglement. We further release USO-Bench, to the best of our knowledge, the first benchmark tailored for evaluating cross-task customization.

\textbf{Performance: }Extensive evaluations on USO-Bench and DreamBench~\cite{ruiz2023dreambooth} show that USO achieves state-of-the-art results on subject-driven, style-driven, and joint style-subject-driven tasks, attaining the highest CLIP-T, DINO, and CSD scores. USO can handle individual tasks and their free-form combinations while exhibiting superior subject consistency, style fidelity, and text controllability as shown in ~\Cref{teaser}.

% style数据+subject模型 => style triplet data
% style triplet data 训练 style+subject模型
% 用style loss监督subject的训练
% 用subject模型构造style数据

% We would even argue that the pre-trained model should be trained on "bad" data, as long as the undesirable aspects of the data are accurately captured in its conditioning. Indeed, in addition to telling the model what we want, we often want to tell it what we don't want.
\section{Related Work}
\label{sec:Related Work}

\subsection{Style Transfer}

Style Transfer aims to apply the style in the reference image to the given content image or fully generated image.
Early work like adaptive instance normalization~\cite{huang2017adain} achieved impressive style transfer results with layout-preserved results by simply using a pre-trained network as the style encoder and well-designed injection modules.

The recent powerful text-to-image generation base models, like Stable Diffusion~\cite{podell2024sdxl, esser2024scaling} and FLUX~\cite{blackforestlabs_flux}, along with style transfer plugins built upon them, have significantly improved the convenience and effectiveness of performing this task.
Several are even training-free, like StyleAlign~\cite{wu2021stylealign} and StylePrompt~\cite{jeong2024visual} which transfer the style via simple query-key swapping in the specific self-attention layers.
Other training-based methods can theoretically achieve better fitting and style transfer performance, but they also raise concerns of content leakage. 
IP-adapter~\cite{ye2023ip} and DEADiff~\cite{qi2024deadiff} demonstrate the style transfer ability with a new decoupled cross-attention layer trained with coupled data, and overcome the content leakage by decreasing the injection weights in inference-time.
InstanceStyle~\cite{wang2024instantstyle}, StyleShot~\cite{gao2024styleshot} and B-lora~\cite{frenkel2024implicit} provide more detailed time-aware and layer-aware injection strategies to disentangle the style and content feature injections.
However, those disentanglement analyses are tied to the specific model architecture and hard to migrate.

\subsection{Subject-Driven generation}

Subject-driven generation refers to generating images of the same subject conditioned on a text instruction and reference images of given subjects.
Dreambooth~\cite{ruiz2023dreambooth} and IP-Adapter~\cite{ye2023ip} turn a UNet-based text-to-image model into a subject-driven model by parameter-efficient tuning or a newly introduced attention plug-in.
Recently, popular image-generation foundation models have shifted from UNet-based architectures to transformer-based ones.
The inherent in-context learning capabilities of transformers have greatly enriched research on subject-driven generation. ICLoRA~\cite{huang2024context}, OmniControl~\cite{tan2024ominicontrol}, UNO~\cite{wu2025less}, and FLUX.1 Kontext~\cite{labs2025flux} use shared attention between the generated image and reference image to train a text-to-image DiT into a subject-driven variant.
It is worth noting that some of them have extended the reference subject to other types.
OmniControl~\cite{tan2024ominicontrol} supports layout control image as a reference, UNO~\cite{wu2025less} supports multiple reference images input, and DreamO~\cite{mou2025dreamo} can work for simple style transfer.
They have indicated that various types of reference-guided generation can be unified within the DiT in-context framework.
This further prompts the question of whether jointly addressing different tasks in this setting could lead to mutual benefits across them.

\section{Methodology}
\label{sec:Methodology}
\subsection{Preliminary}
Latent diffusion models~\cite{rombach2022high, podell2024sdxl} have evolved from UNet-based architectures to DiT-based designs, with steadily improving foundational capabilities. MM-DiT~\cite{esser2024scaling, blackforestlabs_flux} further elevates image-generation quality, spawning numerous downstream applications and unlocking greater controllability in text-to-image generation~\cite{wu2025less}. It incorporates a multi-modal attention mechanism that can be seamlessly extended to an in-context generation framework: the conditioned tokens are directly concatenated with the text prompt and the noisy latent, yielding the formulation:
\begin{equation}
\operatorname{Attention}\left([c, z_t, z_c]\right)=\operatorname{softmax}\left(\frac{\mathbf{Q} \mathbf{K}^\top}{\sqrt{d}}\right)\mathbf{V},\label{eq1}
\end{equation}
where $Z=[c, z_t, z_c]$ denotes the concatenation of text tokens, noisy latent, and condition tokens. This allows both representations to function within their own respective spaces while still taking the other into account.

\begin{figure}[t]
\centering
\includegraphics[width=1\textwidth]{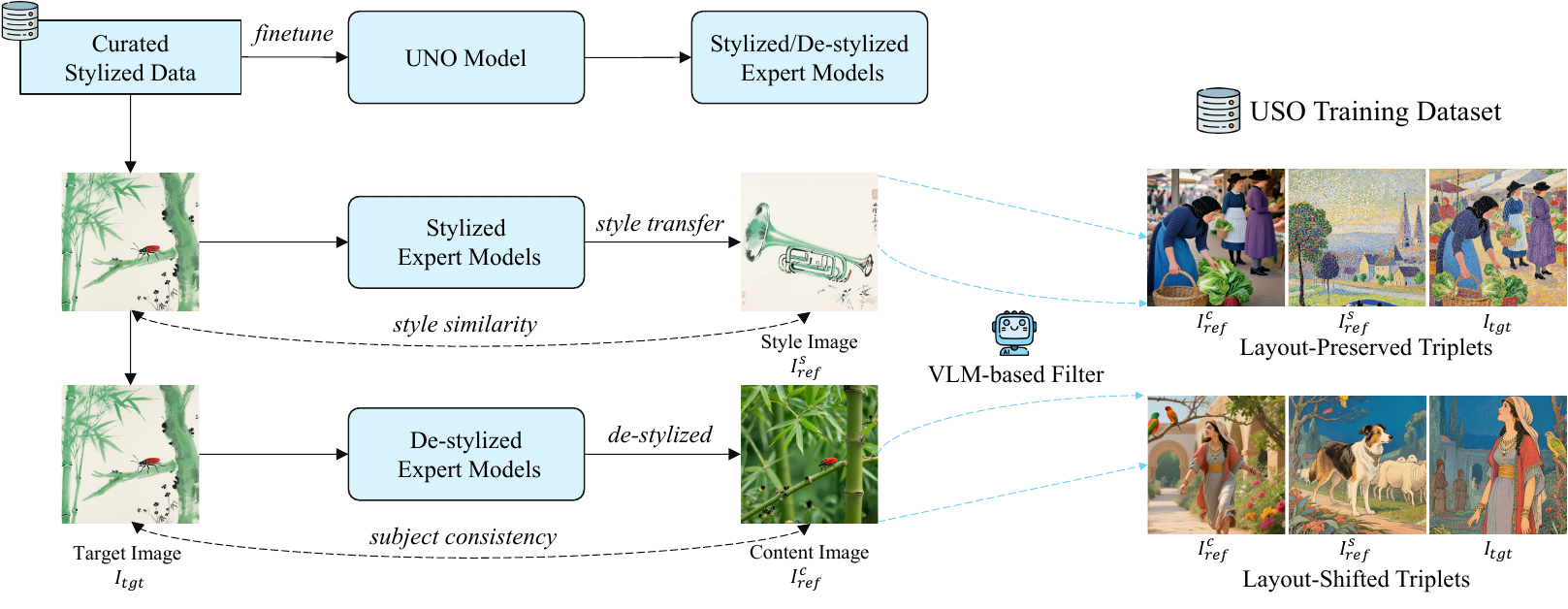}
    \caption{Illustration of our proposed cross-task triplet curation framework, which systematically generates layout-preserved and layout-shifted triplets.}
    \label{fig2}
\end{figure}
\subsection{Cross-Task Triplet Curation Framework}\label{uno_dataset}
This section details the construction of cross-task triplets for USO training. Although prior works~\cite{xing2024csgo,wang2025omnistyle} have explored triplet generation, they retain the original layout, preventing any pose or spatial re-arrangement of the subject. To jointly enable subject-driven and style-driven generation beyond simple instruction-based edits, we curate a new USO dataset expressly designed for this unified objective.

~\Cref{fig2} provides an overview of USO dataset. Our co-disentanglement paradigm starts from a \textit{subject-for-style} data curation framework. Among many possible tasks, subject-driven (i.e., UNO-1M~\cite{wu2025less}) and instruction-based editing (i.e., X2I2~\cite{wu2025omnigen2}) datasets are comparatively easy to collect at scale, enabling targeted task-specific corpora. In particular, subject-driven data emphasizes learning from content cues while preserving subject identity and consistency; instruction-based editing bridges styles by preserving spatial layout and transferring appearance between realistic and stylized domains in both directions. These resources naturally support training domain-specialist models and, through deliberate dataset design, induce the capabilities we care about (\textit{e.g.}, extracting task-relevant features conditioned on image type). Guided by these insights, we curate $200k$ stylized image pairs sourced from publicly licensed datasets and augmented with samples synthesized by state-of-the-art text-to-image models. Using these data, we trained two complementary experts on top of the leading customization framework UNO~\cite{wu2025less}:
\textbf{(1) a stylized expert model} that performs style-driven generation conditioned on a style-reference image, producing a new subject rendered in the target style ($I_{ref}^s$ from $I_{tgt}$), and \textbf{(2) a de-stylization expert model} that inverts a stylized image to a photorealistic counterpart, allowing either flexible layout shifts or preservation ($I_{ref}^c$ from $I_{tgt}$). 

Each curated stylized image serves as the target $I_{tgt}$. We synthesize its style reference $I^s_{ref}$ via the stylization expert and its content reference $I^c_{ref}$ via the de-stylization expert. Following~\cite{wu2025less}, a VLM-based filter enforces style similarity between $I_{tgt}$ and $I^s_{ref}$ and subject consistency between $I_{tgt}$ and $I^c_{ref}$. This yields two kinds of triplets, shown in ~\Cref{fig2}: layout-preserved and layout-shifted. Unlike prior work~\cite{xing2024csgo,wang2025omnistyle}, which focuses solely on style-driven generation and confines itself to layout-preserved triplets, our cross-task triplets achieve deeper content–style disentanglement across tasks and are used to train USO.

\begin{figure*}[t]
\centering
\includegraphics[width=\textwidth]{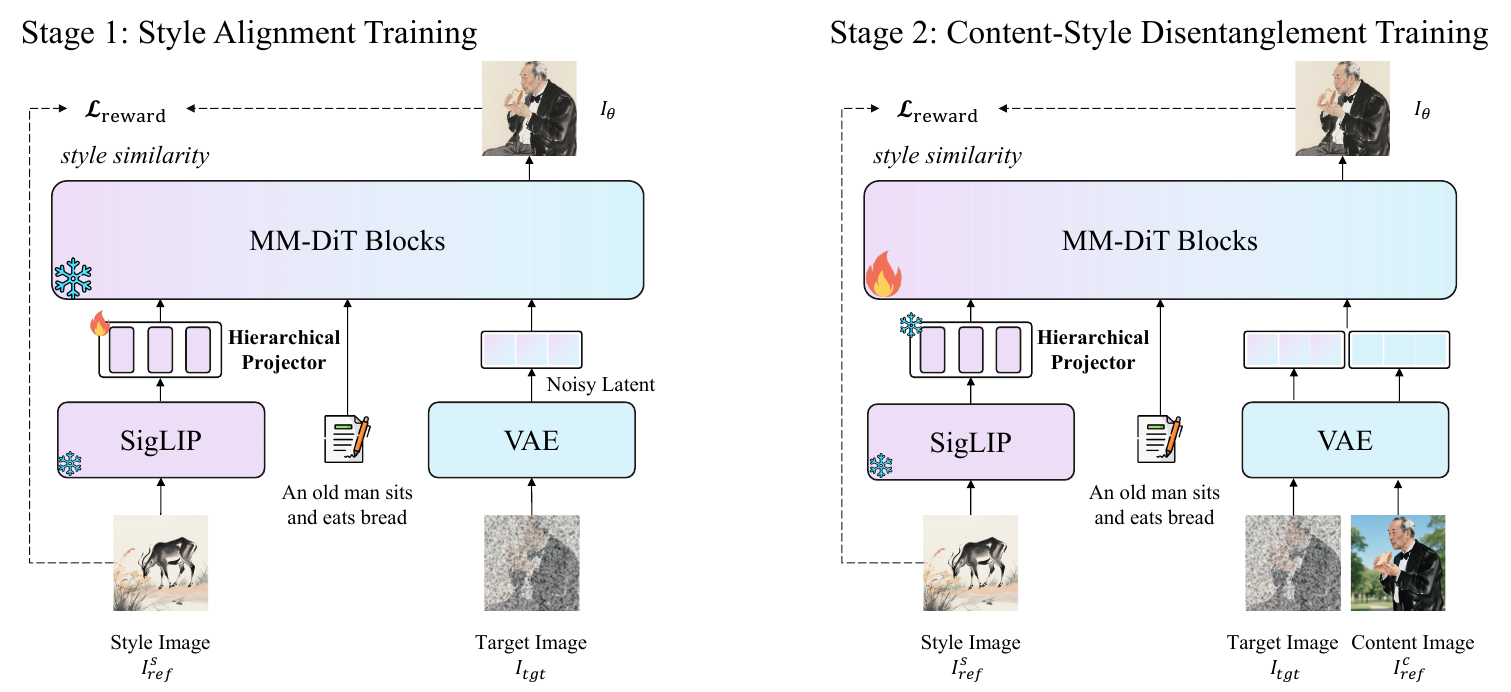}
    \caption{Illustration of the training framework of \textbf{USO}. USO unifies subject-driven and style-driven generation in two stages: Stage 1 aligns SigLIP embeddings via style-alignment training to yield a style-capable model; Stage 2 disentangles the conditional encoders and trains on triplets to enable the joint conditional generation. Finally, a style-reward learning paradigm supervises both stages to yield a stronger unified model.}
    \label{fig3}
\end{figure*}
\subsection{Unified Customization Framework (USO)}
In this section, we describe how we unify two tasks that have traditionally been treated separately, style-driven and subject-driven generation, into a single model. Each task demands the model to master distinct knowledge: the former emphasizes style similarity, while the latter insists on subject consistency. By excelling at both simultaneously, the model naturally disentangles content from style, a long-standing focus of style-driven generation, which in turn boosts the quality of stylization and customization. Beyond merely preserving layout during style-driven generation, the model can now freely recombine any subject with any style. 

\subsubsection{Style Alignment Training.}
As illustrated in ~\Cref{fig3}, We start from a pre-trained text-to-image (T2I) model and fine-tune it into a stylized variant via our proposed style-alignment training. Unlike prior in-context generation approaches that rely solely on a VAE $\mathcal{E}(\cdot)$ to encode the conditioned image $I_{ref}$, we argue that style is a more abstract cue demanding richer semantic information. Therefore, we employ the semantic encoder SigLIP instead of the VAE to process the reference style image $I^s_{ref}$. While subject-driven or identity-preserving tasks typically emphasize high-level semantics, style-driven tasks must simultaneously handle two extremes: high-level semantics to accommodate large geometric deformations (e.g., 3-D cartoon styles) and low-level details to reproduce subtle brushstrokes (e.g., pencil sketches). Following recent works like~\cite{zhang2024ssr}, we introduce a lightweight Hierarchical Projector $\mathcal{M}_{\text{Proj}}(\cdot)$ to project multi-scale, fine-grained visual features $z_s$ from the extracted SigLIP embeddings $\{c_i\}_{i=1}^N$, where $N$ represents the layer indices of SigLIP. This process can be formulated as:
\begin{equation}
z_s = \operatorname{Concatenate}(\mathcal{M}_{\text{Proj}}(\{c_i\}_{i=1}^N)),\label{eq2}
\end{equation}
Specifically, we assign the style tokens $z_s$ the same positional indices as the text tokens $c$, then feed the fused multimodal sequence $z_1$ into the DiT model as its input:
\begin{equation}
z_1 = \operatorname{Concatenate}(z_s, c, z_t),\label{eq3}
\end{equation}

In this stage, we freeze all parameters except the Hierarchical Projector, enabling the extracted style features to be rapidly aligned with the native textual distribution. Consequently, the pretrained T2I model is converted into a stylized variant capable of accepting style-reference images as conditional input.

\subsubsection{Content-Style Disentanglement Training.}
Building on Stage 1, we introduce subject conditioning in Stage 2 as shown in \Cref{fig3}. Following recent paradigms~\cite{tan2024ominicontrol,wu2025less}, the content image $I^c_{ref}$ is encoded into pure conditional tokens $z_c$ by a frozen VAE encoder $\mathcal{E}(\cdot)$. We formulate USO as a multi-image conditioned model, yet explicitly disentangle content and style features via separate encoders. This design alleviates content leakage, where extraneous style-image details undesirably appear in the output, and also helps the model learn to exclude undesired features for the specific task, as introduced in \Cref{sec:intro}.

During training, the Hierarchical Projector remains frozen while the DiT parameters are unfrozen. Content tokens receive positional indices via UnoPE~\cite{wu2025less} in its diagonal layout. The final multimodal input sequence $z_2$ is thus expressed as:
\begin{equation}
z_2 = \operatorname{Concatenate}(z_s, c, z_t, z_c),\label{eq3}
\end{equation}
Consequently, USO can directly handle both subject-driven and style-driven tasks on the proposed triplet dataset.

Compared with prior open-source style-driven methods, most of which either (i) rigidly preserve the content layout while altering its style~\cite{wang2025omnistyle} or (ii) retain layout via an external ControlNet at the cost of subject consistency with the content image~\cite{qi2024deadiff, wang2024instantstyle}-USO removes these constraints. Trained on our triplet data, it freely re-positions the subject from the content image into any scene while re-rendering it in the style of the reference image.

\subsubsection{Style Reward Learning}
Beyond the standard flow-matching objective, we introduce Style Reward Learning (SRL) to explicitly disentangle style from content during optimization. Flow-matching pre-trains the model by minimizing the L2 distance between the predicted velocity $\bm{v}_\theta(\bm{x}_t,t)$ and the true velocity $\bm{v}_t=\frac{d\alpha_t}{dt}\bm{x}_0+\frac{d\sigma_t}{dt}\epsilon$. Building on this, we denote the training objective as $\mathcal{L}_{\text{Pre}}$, which can be computed as:
\begin{equation}
\mathcal{L}_{\text{Pre}}=\mathbb{E}_{\bm{x}_0,t,\epsilon}[w(t)\|\bm{v}_\theta-\bm{v}_t\|^2]\label{eq5}
\end{equation}
where $w(t)$ is a weighting function, $\bm{v}_\theta$ is a neural network parameterized by $\theta$, and $\alpha_t=1-t, \sigma_t=t$ are continuous-time coefficients with $t\in[0,T=1]$. The sampling process is from $t=T$ with $\bm{x}_T\sim\mathcal{N}(\bm{0}, \bm{I})$ and stops at $t=0$, solving the PF-ODE via $d\bm{x}_t=\bm{v}_\theta(\bm{x}_t,t)dt$.

Extending ReFL~\cite{xu2023imagereward} from T2I generation to the reference-to-image setting, where generation is conditioned on both an image reference and and its corresponding text prompt, SRL alternates between computing a reward score and back-propagating the reward signal. As shown in ~\Cref{fig3}, we define the reward score as the style similarity between the reference style image $I^s_{ref}$ and the generated stylized image $I_{\theta}$, measured by either a VLM-based filter or the CSD model $\mathcal{M}_{\text{RM}}(\cdot)$~\cite{somepalli2024measuring, xing2024csgo}. The reward loss is defined as:
\begin{equation}
    \mathcal{L}_{\text{SRL}}=\mathbb{E}_{y_i\sim \mathcal{Y}}[\phi(\mathcal{M}_{\text{RM}}(y_i,I_\theta(y_i)))]\label{eq6}
\end{equation}
where $\mathcal{Y}=\{y_i\}^n_{i=1}$ is the prompt set, $\phi$ maps reward scores to per-sample loss values, and $I_\theta$ denotes the image generated by the diffusion model with parameters $\theta$ corresponding to prompt $y$.

The final objective combines both losses:
\begin{equation}
\mathcal{L}=\mathcal{L}_{\text{Pre}}+\lambda\mathcal{L}_{\text{SRL}}, \quad \lambda=0\text{ before step }S,\ \lambda=1\text{ thereafter}.
\end{equation}

As shown in ~\Cref{alg:flow_srl}, we present the detailed SRL algorithm. The entire process comprises gradient-free inference followed by a reward-backward step.

\begin{algorithm}[tb]
\caption{Style Reward Learning (SRL) with Flow Matching}
\begin{algorithmic}[1]
\REQUIRE Customization model $\texttt{net}$ with pretrained parameters $\theta$; pretrain loss $\mathcal{L}_{\text{Pre}}$; reward loss $\mathcal{L}_{\text{SRL}}$; reward model $\mathcal{M}_{\text{RM}}$; balancing coefficient $\lambda$; noise-schedule steps $T$; SRL fine-tuning interval $[t_s,t_e]$; dataset $\mathcal{D}=\{(y,I_0,I^c_{\text{ref}},I^s_{\text{ref}})\}$, $y$ is prompt, $I_0$ is target image and $I^c_{ref},I^s_{ref}$ are reference content and style images (\Cref{uno_dataset})

\FOR{$(y,I_0,I^c_{\text{ref}},I^s_{\text{ref}})\in\mathcal{D}$}
    \STATE $\mathcal{L}_{\text{Pre}}\gets \texttt{net}_{\theta}(y,I_0,I^c_{\text{ref}},I^s_{\text{ref}})$ // calculate pretrain loss with \Cref{eq5}
    \STATE $t\sim\mathcal{U}(t_s,t_e)$ // pick a random time step in $[t_s,t_e]$
    \STATE $x_T\sim\mathcal{N}(\mathbf{0},\mathbf{I})$
    \FOR{$\tau=T,\dots,t+1$}
        \STATE $\hat{v}_{\tau}\gets\text{no-grad}(\texttt{net}_{\theta}(y,x_{\tau},I^c_{\text{ref}},I^s_{\text{ref}}))$
        \STATE $x_{\tau-1}\gets x_{\tau}-\hat{v}_{\tau}\Delta t$ // reverse-step update
    \ENDFOR
    \STATE $\hat{v}_{t}\gets\texttt{net}_{\theta}(y,x_{t},I^c_{\text{ref}},I^s_{\text{ref}})$
    \STATE $\hat{I}_0\gets\operatorname{decode}(x_{t}-\hat{v}_t\Delta t)$ // predict original image
    \STATE $\mathcal{L}_{\text{SRL}}\gets -\mathcal{M}_{\text{RM}}(\hat{I}_0,I^s_{\text{ref}})$ // calculate SRL loss with negative reward with \Cref{eq6}
    \STATE $\mathcal{L}\gets\mathcal{L}_{\text{Pre}}+\lambda\mathcal{L}_{\text{SRL}}$
    \STATE $\theta\gets\theta-\eta\,\nabla_{\theta}\mathcal{L}$ // update model parameters via gradient descent ($\eta$ is learning rate)
\ENDFOR
\end{algorithmic}\label{alg:flow_srl}
\end{algorithm}
\begin{figure*}[ht]
\centering
\vspace*{-0.1\textwidth}
\hspace*{-0.105\textwidth}
\includegraphics[width=1.2\textwidth]{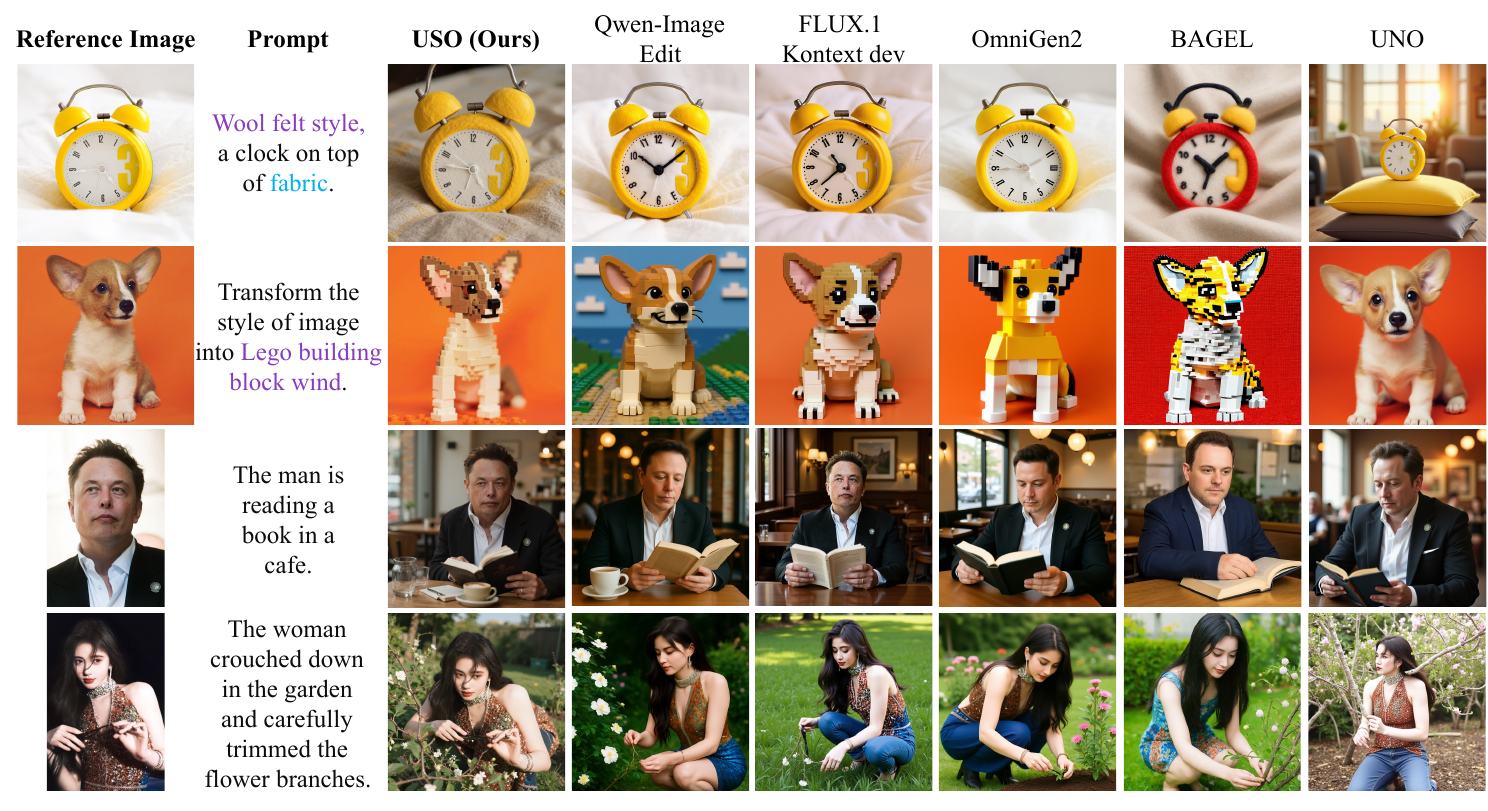}
    \caption{Qualitative comparison with different methods on subject-driven generation.}
    \label{fig4:ip_comparison}
\end{figure*}

\begin{figure*}[h]
\centering
\vspace*{-0.1\textwidth}
\hspace*{-0.1\textwidth}
\includegraphics[width=1.2\textwidth]{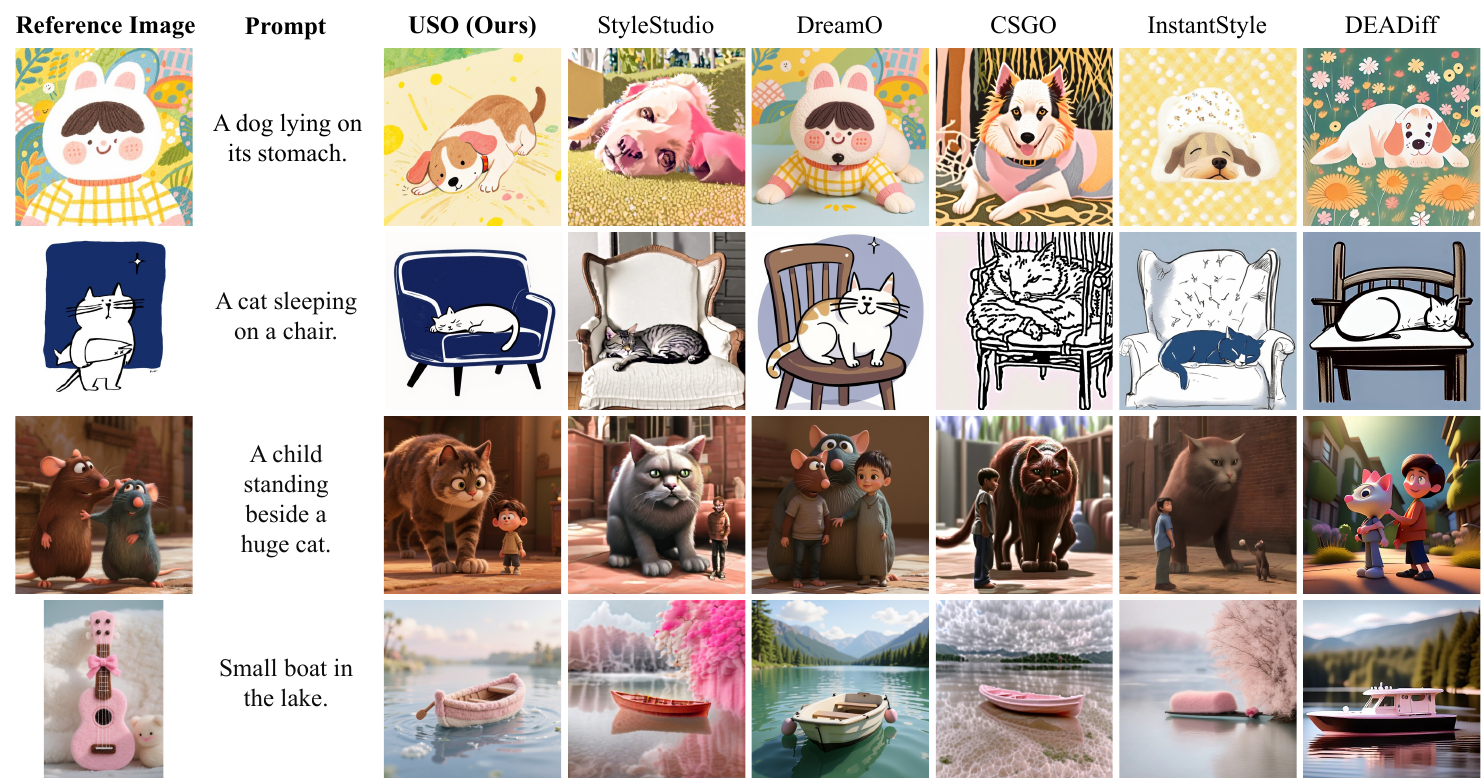}
    \caption{Qualitative comparison with different methods on style-driven generation.}
    \label{fig5:style_comparison}
\end{figure*}
\FloatBarrier

\begin{figure*}[!t]
\centering
\vspace*{-0.12\textwidth}
\includegraphics[width=0.88\textwidth]{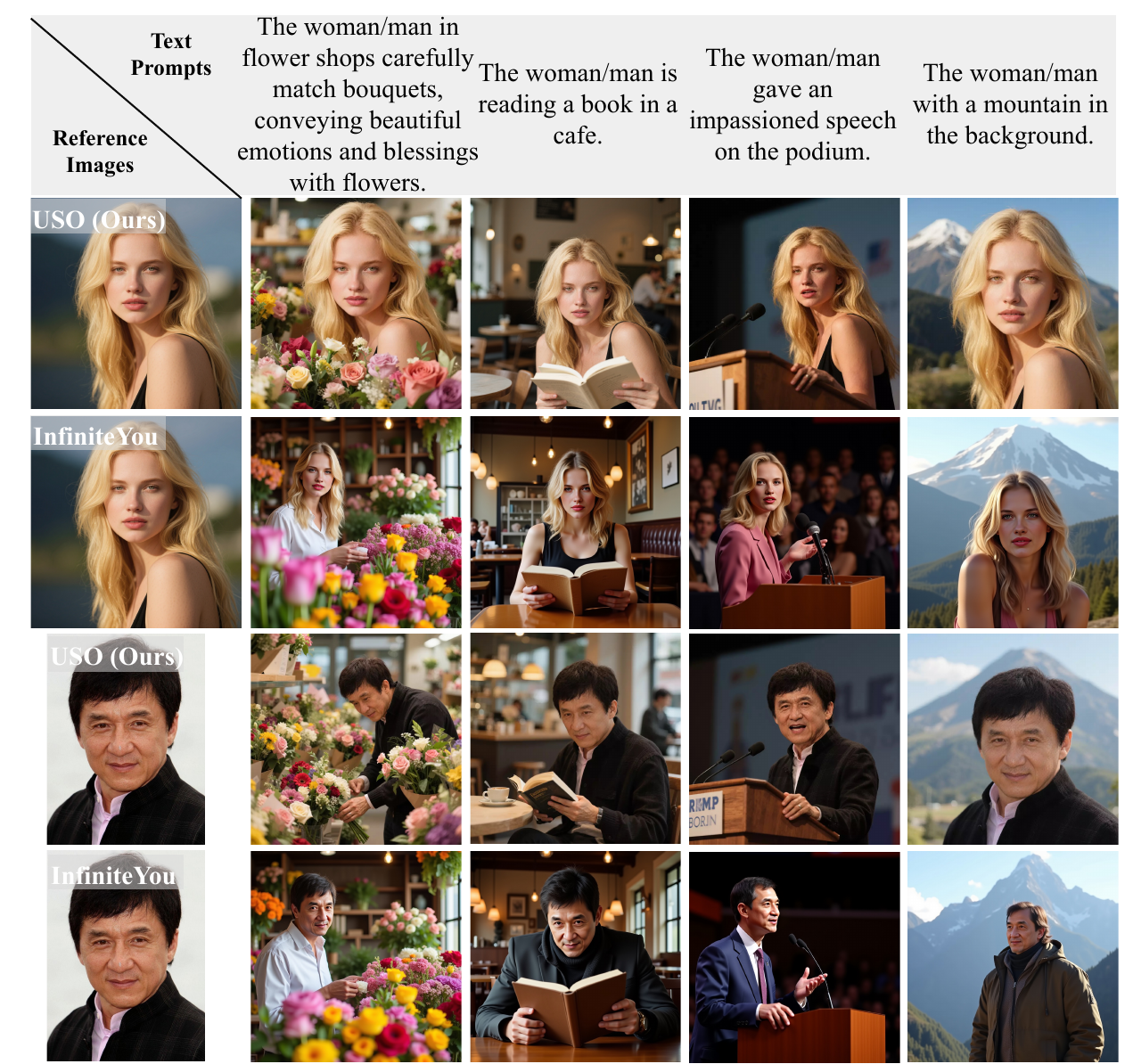}
    \caption{Qualitative comparison with different methods on identity-driven generation.}
    \label{fig4_2:id_comparison}
\end{figure*}

\begin{figure}[h]
\centering
\vspace*{0.04\textwidth}
\hspace*{-0.1\textwidth}
\includegraphics[width=1.2\textwidth]{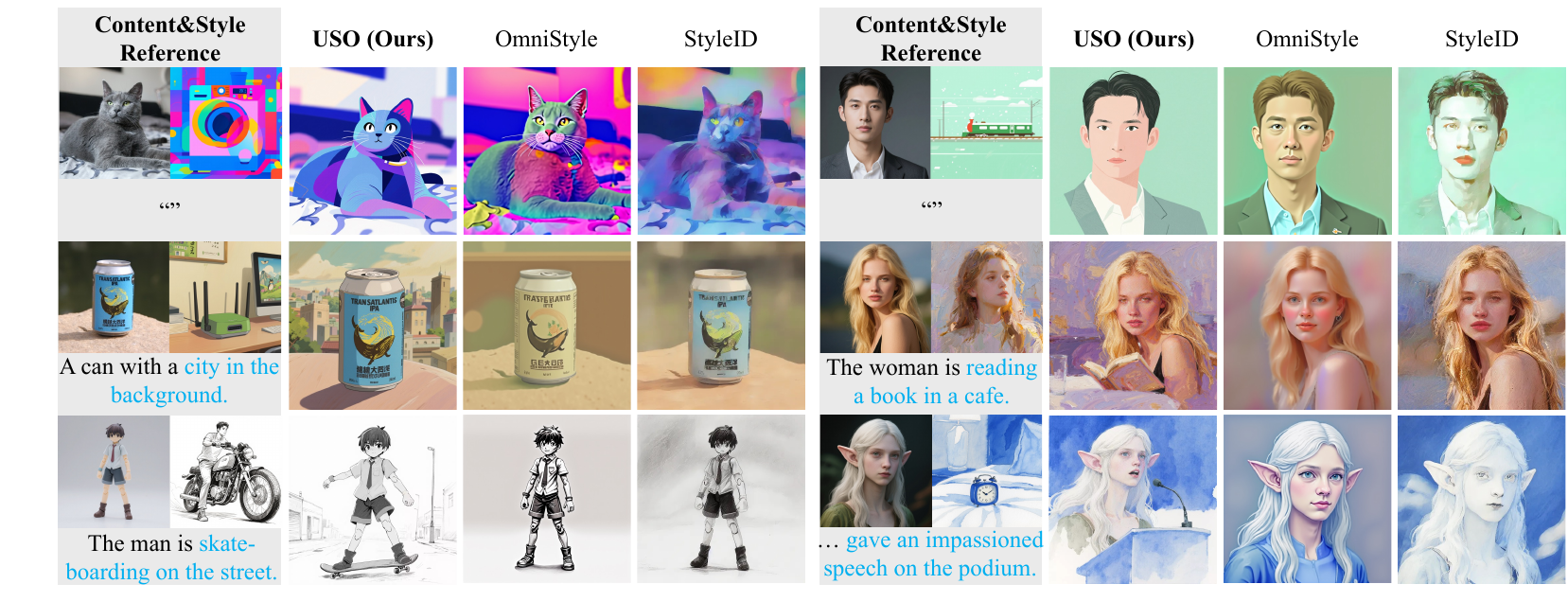}
    \caption{Qualitative comparison with different methods on style-subject-driven generation.\newline}
    \label{fig6:ipstyle_comparison}
\end{figure}

\FloatBarrier

\section{Experiments}
\label{sec:Experiments}
\subsection{Experiments Setting}
% \subsubsection{{Implementation Details.}} % 附加材料
\noindent \textbf{USO Unified Benchmark. }To enable a comprehensive evaluation, we introduce USO-Bench, a unified benchmark built from $50$ content images ($20$ human-centric, $30$ object-centric) paired with $50$ style references. We further craft $30$ subject-driven prompts that span pose variation, descriptive stylization, and instructive stylization, along with $30$ style-driven prompts. We generate four images per prompt for both subject-driven and style-driven tasks, and a single image for the combined style-subject-driven task. This yields $6000$ samples for subject-driven generation, $7040$ for style-driven generation, and $29500$ for the combined task; full construction details are provided in the supplementary material.

\begin{table*}[h]
    \centering
    % \resizebox{0.99\linewidth}{31.5mm}{
    \resizebox{0.86\linewidth}{!}{
        \begin{tabular}{lccc|cc}
        \toprule
        \textbf{Model} & \multicolumn{3}{c|}{\textbf{Subject-driven generation}} & \multicolumn{2}{c}{\textbf{Style-driven generation}}\\
        
        % \cmidrule(lr){3-6} \cmidrule(lr){7-8} \cmidrule(lr){9-10} \cmidrule(lr){11-15}
        &CLIP-I$\uparrow$ & DINO$\uparrow$ & CLIP-T$\uparrow$ & CSD$\uparrow$ & CLIP-T$\uparrow$  \\
        \midrule
        % \multicolumn{14}{l}{\textit{Image Understanding}} \\
        RealCustom++~\cite{huang2024realcustom} & 0.314 & 0.615 & \textbf{0.303} & - & - \\
        RealGeneral~\cite{lin2025realgeneral} & 0.485 & 0.732 & 0.275  & - & -\\
        UNO~\cite{wu2025less} & \underline{0.605} & \underline{0.789} & 0.264  & - & -\\
        BAGEL~\cite{deng2025emerging}  & 0.516 & 0.741 & 0.298 & - & -\\ 
        OmniGen2~\cite{wu2025omnigen2} & 0.475 & 0.723 & \underline{0.302}  & - & -\\
        FLUX.1 Kontext dev~\cite{labs2025flux} & 0.579 & 0.775 & 0.287  & - & -\\
        Qwen-Image Edit~\cite{wu2025qwen} & 0.544 & 0.756 & \underline{0.302}  & - & -\\
        \midrule
        % \multicolumn{14}{l}{\textit{Text-to-Image Generation}} \\
        DEADiff~\cite{qi2024deadiff} & - & - & - & 0.462 & 0.274 \\
        InstantStyle-XL~\cite{wang2024instantstyle} & - & - & - & \underline{0.540} & 0.276 \\
        CSGO~\cite{xing2024csgo} & - & - & - & 0.452 & 0.272 \\
        StyleStudio~\cite{lei2025stylestudio}  & - & - & - & 0.348 & \textbf{0.282} \\
        \midrule
        DreamO~\cite{mou2025dreamo} & 0.588 & 0.787 & 0.280  & 0.454 & \underline{0.278} \\
        \midrule
        \rowcolor{myblue} \textbf{USO (Ours)}  & \textbf{0.623} & \textbf{0.793} & 0.288 & \textbf{0.557} & \textbf{0.282} \\

        \bottomrule
        \end{tabular}
}
     \caption{Quantitative results for subject-driven and style-driven generation on USO-Bench.}
    \label{tab:main_comparison}
\end{table*}
\begin{table}[h]
    \centering
    % \resizebox{0.99\linewidth}{31.5mm}{
    \resizebox{0.38\linewidth}{!}{
        \begin{tabular}{lcc}
        \toprule
        \textbf{Model} & CSD$\uparrow$ & CLIP-T$\uparrow$ \\
        \midrule
        StyleID~\cite{chung2024style} & \underline{0.407} & \underline{0.230} \\
        OmniStyle~\cite{wang2025omnistyle} & 0.365 & 0.229 \\
        \midrule
        \rowcolor{myblue} \textbf{USO (Ours)}  & \textbf{0.495} & \textbf{0.283} \\
        \bottomrule
        \end{tabular}
}
     \caption{Quantitative results for style-subject-driven driven generation on USO-Bench.}\label{tab:ip_style_comparison}
\end{table}
\noindent \textbf{Evaluation Metrics. }For quantitative evaluation, we assess each task along three dimensions: \textbf{(1) subject consistency}, measured by the cosine similarity of CLIP-I and DINO embeddings following \cite{wu2025less}; \textbf{(2) style similarity}, reported via the CSD score \cite{somepalli2024measuring} for both style-driven and style-subject-driven generation, following \cite{xing2024csgo}; and \textbf{(3) text–image alignment}, evaluated with CLIP-T across all three tasks.

\noindent \textbf{Comparative Methods. }As a unified customization framework, USO is evaluated against both task-specific and unified baselines. For subject-driven generation, we benchmark RealCustom++~\cite{mao2024realcustom++}, RealGeneral~\cite{lin2025realgeneral}, UNO~\cite{wu2025less}, OmniGen2~\cite{wu2025omnigen2}, BAGEL~\cite{deng2025emerging}, FLUX.1 Kontext dev~\cite{labs2025flux}, and Qwen-Image Edit~\cite{wu2025qwen}. For style-driven generation, we compare StyleStudio~\cite{lei2025stylestudio}, DreamO~\cite{mou2025dreamo}, CSGO~\cite{xing2024csgo}, InstantStyle~\cite{wang2024instantstyle}, and DEADiff~\cite{qi2024deadiff}. For the joint style-subject-driven setting with dual conditioning, we compare OmniStyle~\cite{wang2025omnistyle} and StyleID~\cite{chung2024style}. We also compared with InfiniteYou~\cite{jiang2025infiniteyou} to further demonstrate the positive effect of our proposed method on identity tasks.

\subsection{Experimental Results}
\noindent \textbf{Subject-Driven Generation. }As shown in ~\Cref{fig4:ip_comparison}, the first two columns demonstrate that USO simultaneously satisfies both descriptive and instructive style edits while maintaining high subject consistency. In contrast, competing methods either fail to apply the style or lose the subject. The last two columns further illustrate USO’s strength in preserving human appearance and identity; it adheres strictly to the textual prompt and almost perfectly retains facial and bodily features, whereas other approaches fall short. When the prompt is “The man is reading a book in a cafe”, FLUX.1 Kontext dev~\cite{labs2025flux} achieves decent facial similarity but carries copy-paste risks. In~\Cref{fig4_2:id_comparison} we compare with task-specific identity-preserving methods; USO produces more realistic, non-plastic results with higher identity consistency. As reported in~\Cref{tab:main_comparison}, USO significantly outperforms prior work, achieving the highest DINO and CLIP-I scores and a leading CLIP-T score.

\noindent \textbf{Style-Driven Generation. }~\Cref{fig5:style_comparison} shows that USO outperforms task-specific baselines in preserving the original style, including global color palettes and painterly brushwork. In the last two columns, given highly abstract references such as material textures or Pixar-style renderings, USO handles them almost flawlessly while prior methods struggle, demonstrating the generalization power of our cross-task co-disentanglement. Quantitatively, ~\Cref{tab:main_comparison} confirms that USO achieves the highest CSD and CLIP-T scores among all style-driven approaches.

\noindent \textbf{Style-Subject-Driven Generation. }As illustrated in ~\Cref{fig6:ipstyle_comparison}, we evaluate USO on both layout-preserved and layout-shifted scenarios. When the input prompt is empty, USO not only preserves the original layout of the content reference but also delivers the strongest style adherence. In the last two columns, under a more complex prompt, USO simultaneously preserves the subject and identity consistency, matches the reference style, and aligns with the text, while other methods lag markedly and merely adhere to the text. \Cref{tab:ip_style_comparison} corroborates these observations, showing USO achieves the highest CSD and CLIP-T scores and substantially outperforms all baselines.

\begin{figure*}[h]
\centering
\includegraphics[width=\textwidth]{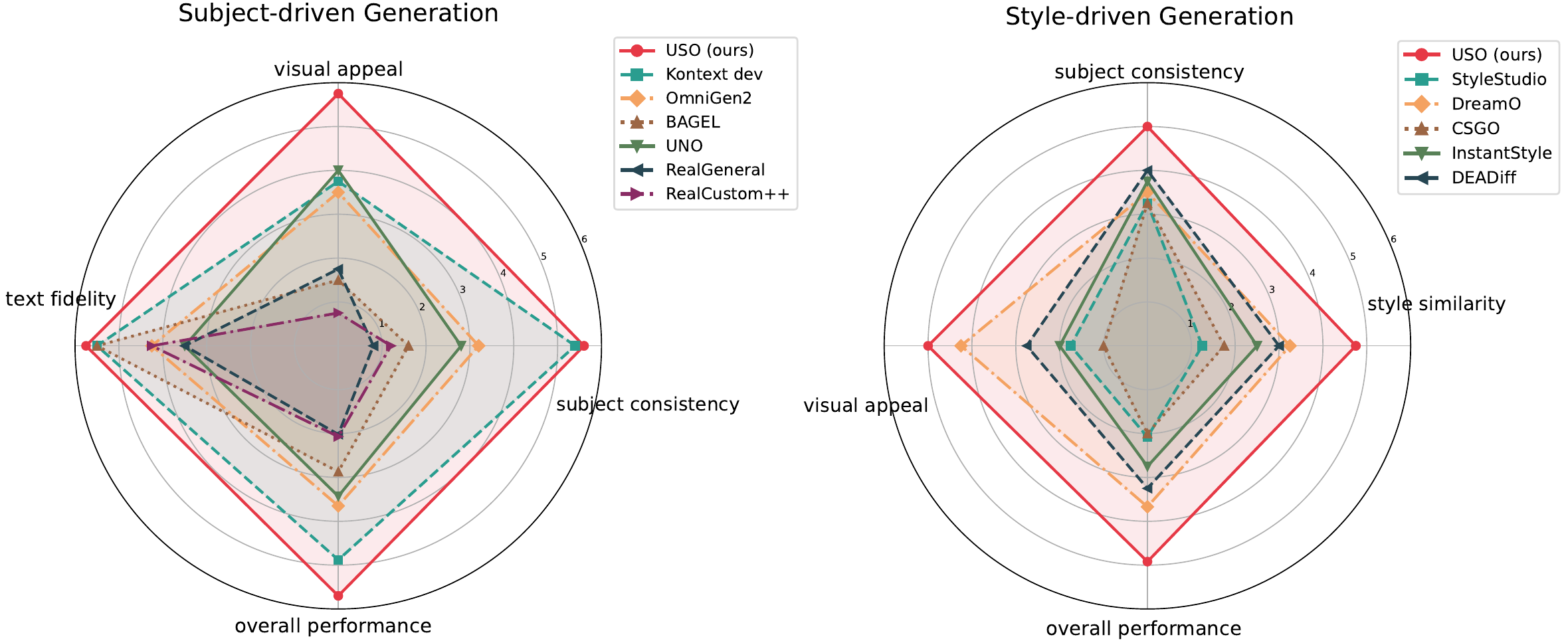}
    \caption{Radar charts of user evaluation of methods for subject-driven and style-driven generation on different dimensions.}
    \label{sup2:user_study}
\end{figure*}

\noindent \textbf{User Study. }We further conduct an online user-study questionnaire to compare state-of-the-art subject-driven and style-driven methods. Questionnaires were distributed to both domain experts and non-experts, who ranked the best results for each task. (1) \textit{Subject-driven tasks} were evaluated on text fidelity, visual appeal, subject consistency, and overall quality. (2) \textit{Style-driven tasks} were judged on text fidelity, visual appeal, style similarity, and overall quality.
As shown in ~\Cref{sup2:user_study}, our USO achieves top performance on both tasks, validating the effectiveness of our cross-task co-disentanglement and showcasing its capability to deliver state-of-the-art results.

\newpage
\section{Ablation Study}

\subsection{Effect of Style Reward Learning (SRL). }
\begin{figure*}[h]
\centering
\includegraphics[width=\textwidth]{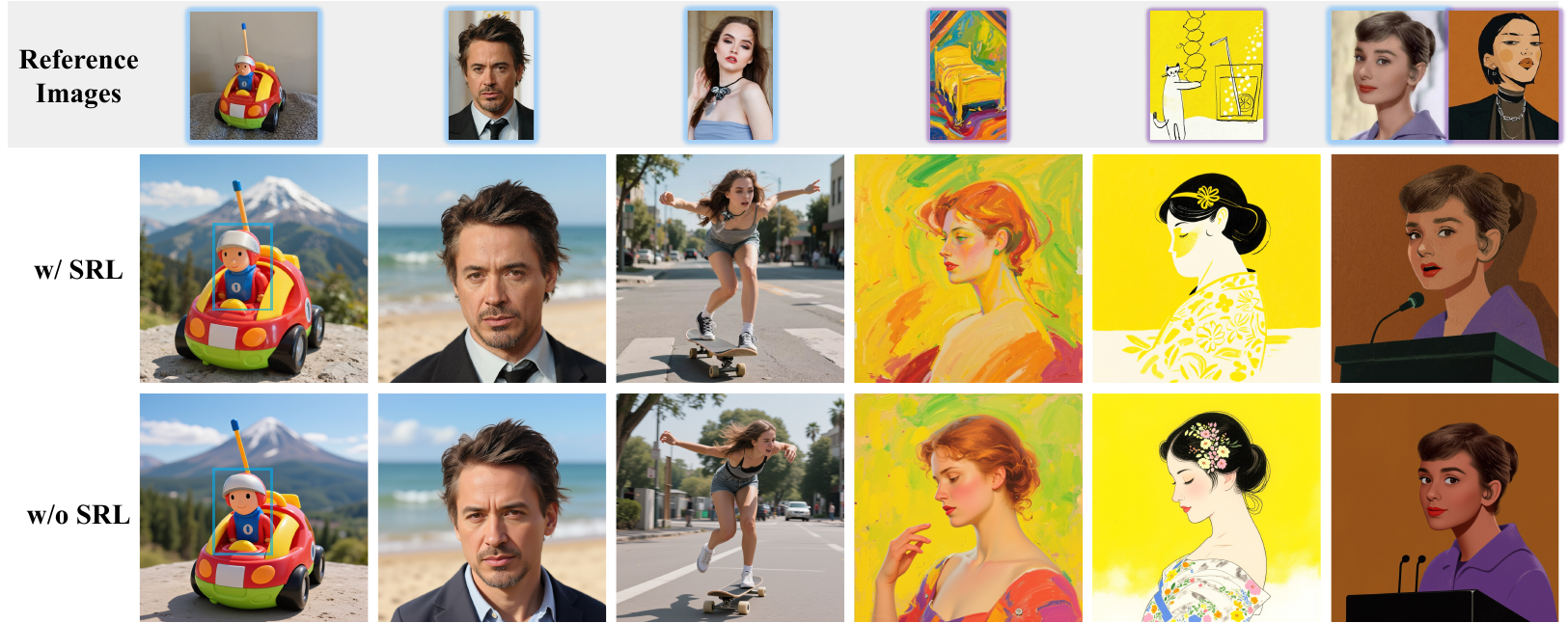}
    \caption{Ablation study of SRL. The \textcolor{mytextblue}{blue} boxes denote content reference and the \textcolor{mypurple}{purple} boxes denote style reference. Prompts are "A toy with a mountain in the background.", "The man on the beach.", "The woman is skateboarding on the street.", "A beautiful woman.", "A beautiful woman.", "The woman gave an impassioned speech on the podium." from left to right.}
    \label{fig7_2:comparison_ab_srl}
\end{figure*}

\noindent \textbf{For style-driven task:} As shown in~\Cref{fig7_2:comparison_ab_srl}, the last three columns reveal a clear boost in style similarity for both style-driven and style-subject-driven tasks; the stroke textures and painting style closely match the reference images, confirming the effectiveness of our style reward learning.

\noindent \textbf{For subject-driven task:} In the first three and final columns of~\Cref{fig7_2:comparison_ab_srl}, we observe a notable improvement in subject and identity consistency, with more uniform details and higher facial similarity.

As shown in~\Cref{tab:main_ab}, removing SRL leads to a sharp drop in the CSD score and simultaneous declines in CLIP-I and CLIP-T. Notably, we rely solely on style reward and introduce no identity-specific data; nevertheless, the unified model benefits in content consistency. By sharpening the model’s ability to extract and retain desired features, SRL yields an overall improvement across all tasks, strongly validating our motivation. Beyond gains in subject and identity fidelity, we observe a noticeable enhancement in aesthetic quality (e.g., texture as in VMix~\cite{wu2024vmix}) and a marked reduction in the plastic artifact, which is an issue that has long challenged text-to-image generation~\cite{wu2024vmix}. Through SRL training, the model exhibits emerging properties even in tasks not explicitly targeted during training.

\begin{figure}[thb]
\CenterFloatBoxes
\begin{floatrow}

% Left Table
\ttabbox
{
\resizebox{\linewidth}{!}{
    \begin{tabular}{lcc|cc}
    \toprule
    \textbf{Model} & \multicolumn{2}{c|}{\textbf{Subject-driven}} & \multicolumn{2}{c}{\textbf{Style-subject-driven}} \\
    
    % \cmidrule(lr){3-6} \cmidrule(lr){7-8} \cmidrule(lr){9-10} \cmidrule(lr){11-15}
    &CLIP-I$\uparrow$ & CLIP-T$\uparrow$ & CSD$\uparrow$ & CLIP-T$\uparrow$ \\
    \midrule
    \textbf{USO (Ours)} & \textbf{0.623} & \textbf{0.288} & \textbf{0.495} & \textbf{0.283}  \\
    \midrule
    w/o SRL & 0.620 & \underline{0.284}  & \underline{0.413} & \underline{0.280} \\
    w/o SAT & \underline{0.621} & 0.275 & 0.409 & \underline{0.280}\\
    w/o DE & 0.594 & 0.269 & 0.382 & 0.277 \\
    \bottomrule
    \end{tabular}
}}
{\caption{Ablation study of different components proposed in USO. }
 \label{tab:main_ab}
}

% Right Table
\ttabbox
{
\resizebox{0.965\linewidth}{!}{
    \begin{tabular}{lcc}
    \toprule
    \textbf{Model} & CSD$\uparrow$ & CLIP-T$\uparrow$ \\
    \midrule

    resampler (depth=1) & \underline{0.336} & 0.279 \\
    resampler, unfreeze siglip & 0.155 & \textbf{0.288} \\
    mlp (depth=1) & 0.277 & \underline{0.284} \\
    mlp, unfreeze siglip & 0.179 & \textbf{0.288} \\
    hierarchical projector & \textbf{0.402} & \underline{0.284} \\

    \bottomrule
    \end{tabular}
}}
{\caption{Ablation study of different projector in USO.}
 \label{tab:stage1_ab}
}

\end{floatrow}
\end{figure}

\subsection{Effect of Style Alignment Training (SAT).}
\begin{figure}[t]
\centering
\includegraphics[width=0.85\textwidth]{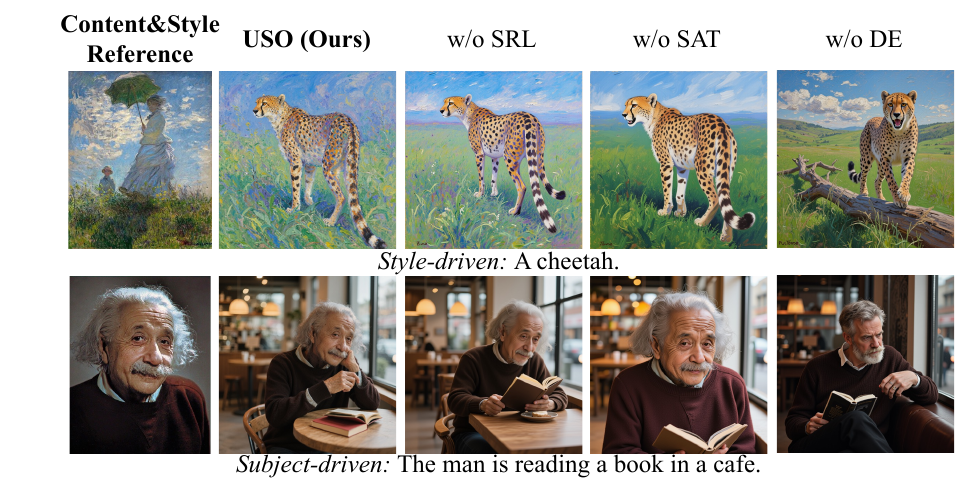}
    \caption{Ablation study of USO. Zoom in for details.}
    \label{fig7:comparison_ab}
\end{figure}

Removing SAT and instead jointly fine-tuning both SigLIP and DiT from scratch degrades CLIP-T on subject-driven tasks and lowers CSD on style-subject-driven tasks (~\Cref{tab:main_ab}). Qualitatively, ~\Cref{fig7:comparison_ab} shows that the oil-painting style of the “cheetah” example becomes noticeably weaker.

\subsection{Effect of Disentangled Encoder (DE).}
Replacing the disentangled encoders with a single VAE for both style and content images harms nearly every metric (~\Cref{tab:main_ab}). Visually, the “cheetah” reverts to a more photorealistic appearance, while the man’s identity suffers a marked loss (~\Cref{fig7:comparison_ab}).

\subsection{Effect of Hierarchical Projector.} ~\Cref{tab:stage1_ab} shows that the hierarchical projector yields the highest CSD and a leading CLIP-T score, substantially benefiting style-alignment training.

\section{Conclusion}
\label{sec:Conclusion}
In this paper, we present USO, a unified framework capable of subject-driven, style-driven, and joint style-subject-driven generation. We introduce a cross-task co-disentanglement paradigm that first constructs a systematic triplet-curation pipeline, then applies progressive style-alignment and content–style disentanglement training on the curated triplets. Additionally, we propose a style-reward learning paradigm to further boost performance. To comprehensively evaluate our method, we construct USO-Bench, a unified benchmark that provides both task-specific and joint evaluation for existing approaches. Finally, extensive experiments demonstrate that USO sets new state-of-the-art results on subject-driven, style-driven, and their joint style-subject-driven tasks, exhibiting superior subject consistency, style fidelity, and text controllability.

\clearpage

\bibliographystyle{plainnat}
\bibliography{main}

\begin{thebibliography}{41}
\providecommand{\natexlab}[1]{#1}
\providecommand{\url}[1]{\texttt{#1}}
\expandafter\ifx\csname urlstyle\endcsname\relax
  \providecommand{\doi}[1]{doi: #1}\else
  \providecommand{\doi}{doi: \begingroup \urlstyle{rm}\Url}\fi

\bibitem[Chen et~al.(2022)Chen, Hu, Saharia, and Cohen]{chen2022re}
Wenhu Chen, Hexiang Hu, Chitwan Saharia, and William~W Cohen.
\newblock Re-imagen: Retrieval-augmented text-to-image generator.
\newblock \emph{arXiv preprint arXiv:2209.14491}, 2022.

\bibitem[Chung et~al.(2024)Chung, Hyun, and Heo]{chung2024style}
Jiwoo Chung, Sangeek Hyun, and Jae-Pil Heo.
\newblock Style injection in diffusion: A training-free approach for adapting large-scale diffusion models for style transfer.
\newblock In \emph{Proceedings of the IEEE/CVF conference on computer vision and pattern recognition}, pages 8795--8805, 2024.

\bibitem[Deng et~al.(2025)Deng, Zhu, Li, Gou, Li, Wang, Zhong, Yu, Nie, Song, et~al.]{deng2025emerging}
Chaorui Deng, Deyao Zhu, Kunchang Li, Chenhui Gou, Feng Li, Zeyu Wang, Shu Zhong, Weihao Yu, Xiaonan Nie, Ziang Song, et~al.
\newblock Emerging properties in unified multimodal pretraining.
\newblock \emph{arXiv preprint arXiv:2505.14683}, 2025.

\bibitem[Esser et~al.(2024)Esser, Kulal, Blattmann, Entezari, M{\"u}ller, Saini, Levi, Lorenz, Sauer, Boesel, et~al.]{esser2024scaling}
Patrick Esser, Sumith Kulal, Andreas Blattmann, Rahim Entezari, Jonas M{\"u}ller, Harry Saini, Yam Levi, Dominik Lorenz, Axel Sauer, Frederic Boesel, et~al.
\newblock Scaling rectified flow transformers for high-resolution image synthesis.
\newblock In \emph{ICML}, 2024.

\bibitem[Frenkel et~al.(2024)Frenkel, Vinker, Shamir, and Cohen-Or]{frenkel2024implicit}
Yarden Frenkel, Yael Vinker, Ariel Shamir, and Daniel Cohen-Or.
\newblock Implicit style-content separation using b-lora.
\newblock In \emph{European Conference on Computer Vision}, pages 181--198. Springer, 2024.

\bibitem[Gal et~al.(2022)Gal, Alaluf, Atzmon, Patashnik, Bermano, Chechik, and Cohen-Or]{gal2022image}
Rinon Gal, Yuval Alaluf, Yuval Atzmon, Or~Patashnik, Amit~H Bermano, Gal Chechik, and Daniel Cohen-Or.
\newblock An image is worth one word: Personalizing text-to-image generation using textual inversion.
\newblock \emph{arXiv preprint arXiv:2208.01618}, 2022.

\bibitem[Gao et~al.(2024)Gao, Liu, Sun, Tang, Zeng, Chen, and Zhao]{gao2024styleshot}
Junyao Gao, Yanchen Liu, Yanan Sun, Yinhao Tang, Yanhong Zeng, Kai Chen, and Cairong Zhao.
\newblock Styleshot: A snapshot on any style.
\newblock \emph{arXiv preprint arXiv:2407.01414}, 2024.

\bibitem[Hu et~al.(2021)Hu, Shen, Wallis, Allen-Zhu, Li, Wang, Wang, and Chen]{hu2021lora}
Edward~J Hu, Yelong Shen, Phillip Wallis, Zeyuan Allen-Zhu, Yuanzhi Li, Shean Wang, Lu~Wang, and Weizhu Chen.
\newblock Lora: Low-rank adaptation of large language models.
\newblock \emph{arXiv preprint arXiv:2106.09685}, 2021.

\bibitem[Huang et~al.(2024{\natexlab{a}})Huang, Wang, Wu, Shi, Dou, Liang, Feng, Liu, and Zhou]{huang2024context}
Lianghua Huang, Wei Wang, Zhi-Fan Wu, Yupeng Shi, Huanzhang Dou, Chen Liang, Yutong Feng, Yu~Liu, and Jingren Zhou.
\newblock In-context lora for diffusion transformers.
\newblock \emph{arXiv preprint arXiv:2410.23775}, 2024{\natexlab{a}}.

\bibitem[Huang et~al.(2024{\natexlab{b}})Huang, Mao, Liu, He, and Zhang]{huang2024realcustom}
Mengqi Huang, Zhendong Mao, Mingcong Liu, Qian He, and Yongdong Zhang.
\newblock Realcustom: narrowing real text word for real-time open-domain text-to-image customization.
\newblock In \emph{CVPR}, pages 7476--7485, 2024{\natexlab{b}}.

\bibitem[Huang and Belongie(2017)]{huang2017adain}
Xun Huang and Serge Belongie.
\newblock Arbitrary style transfer in real-time with adaptive instance normalization.
\newblock In \emph{ICCV}, 2017.

\bibitem[Jeong et~al.(2024)Jeong, Kim, Choi, Lee, and Uh]{jeong2024visual}
Jaeseok Jeong, Junho Kim, Yunjey Choi, Gayoung Lee, and Youngjung Uh.
\newblock Visual style prompting with swapping self-attention.
\newblock \emph{arXiv preprint arXiv:2402.12974}, 2024.

\bibitem[Jiang et~al.(2025)Jiang, Yan, Jia, Liu, Kang, and Lu]{jiang2025infiniteyou}
Liming Jiang, Qing Yan, Yumin Jia, Zichuan Liu, Hao Kang, and Xin Lu.
\newblock Infiniteyou: Flexible photo recrafting while preserving your identity.
\newblock \emph{arXiv preprint arXiv:2503.16418}, 2025.

\bibitem[Labs(2024)]{blackforestlabs_flux}
Black~Forest Labs.
\newblock Flux: Official inference repository for flux.1 models, 2024.
\newblock URL \url{https://github.com/black-forest-labs/flux}.
\newblock Accessed: 2025-02-07.

\bibitem[Labs et~al.(2025)Labs, Batifol, Blattmann, Boesel, Consul, Diagne, Dockhorn, English, English, Esser, et~al.]{labs2025flux}
Black~Forest Labs, Stephen Batifol, Andreas Blattmann, Frederic Boesel, Saksham Consul, Cyril Diagne, Tim Dockhorn, Jack English, Zion English, Patrick Esser, et~al.
\newblock Flux. 1 kontext: Flow matching for in-context image generation and editing in latent space.
\newblock \emph{arXiv preprint arXiv:2506.15742}, 2025.

\bibitem[Lei et~al.(2025)Lei, Song, Zhu, Wang, and Zhang]{lei2025stylestudio}
Mingkun Lei, Xue Song, Beier Zhu, Hao Wang, and Chi Zhang.
\newblock Stylestudio: Text-driven style transfer with selective control of style elements.
\newblock In \emph{Proceedings of the Computer Vision and Pattern Recognition Conference}, pages 23443--23452, 2025.

\bibitem[Li et~al.(2023)Li, Li, and Hoi]{li2023blip}
Dongxu Li, Junnan Li, and Steven Hoi.
\newblock Blip-diffusion: Pre-trained subject representation for controllable text-to-image generation and editing.
\newblock \emph{Advances in Neural Information Processing Systems}, 36:\penalty0 30146--30166, 2023.

\bibitem[Lin et~al.(2025)Lin, Huang, Zhuang, and Mao]{lin2025realgeneral}
Yijing Lin, Mengqi Huang, Shuhan Zhuang, and Zhendong Mao.
\newblock Realgeneral: Unifying visual generation via temporal in-context learning with video models.
\newblock \emph{arXiv preprint arXiv:2503.10406}, 2025.

\bibitem[Mao et~al.(2024)Mao, Huang, Ding, Liu, He, and Zhang]{mao2024realcustom++}
Zhendong Mao, Mengqi Huang, Fei Ding, Mingcong Liu, Qian He, and Yongdong Zhang.
\newblock Realcustom++: Representing images as real-word for real-time customization.
\newblock \emph{arXiv preprint arXiv:2408.09744}, 2024.

\bibitem[Mou et~al.(2025)Mou, Wu, Wu, Guo, Zhang, Cheng, Luo, Ding, Zhang, Li, et~al.]{mou2025dreamo}
Chong Mou, Yanze Wu, Wenxu Wu, Zinan Guo, Pengze Zhang, Yufeng Cheng, Yiming Luo, Fei Ding, Shiwen Zhang, Xinghui Li, et~al.
\newblock Dreamo: A unified framework for image customization.
\newblock \emph{arXiv preprint arXiv:2504.16915}, 2025.

\bibitem[Podell et~al.(2024)Podell, English, Lacey, Blattmann, Dockhorn, M{\"u}ller, Penna, and Rombach]{podell2024sdxl}
Dustin Podell, Zion English, Kyle Lacey, Andreas Blattmann, Tim Dockhorn, Jonas M{\"u}ller, Joe Penna, and Robin Rombach.
\newblock {SDXL}: Improving latent diffusion models for high-resolution image synthesis.
\newblock In \emph{ICLR}, 2024.
\newblock URL \url{https://openreview.net/forum?id=di52zR8xgf}.

\bibitem[Purushwalkam et~al.(2024)Purushwalkam, Gokul, Joty, and Naik]{purushwalkam2024bootpig}
Senthil Purushwalkam, Akash Gokul, Shafiq Joty, and Nikhil Naik.
\newblock Bootpig: Bootstrapping zero-shot personalized image generation capabilities in pretrained diffusion models.
\newblock \emph{arXiv preprint arXiv:2401.13974}, 2024.

\bibitem[Qi et~al.(2024)Qi, Fang, Wu, Xie, Liu, Chen, He, and Zhang]{qi2024deadiff}
Tianhao Qi, Shancheng Fang, Yanze Wu, Hongtao Xie, Jiawei Liu, Lang Chen, Qian He, and Yongdong Zhang.
\newblock Deadiff: An efficient stylization diffusion model with disentangled representations.
\newblock In \emph{Proceedings of the IEEE/CVF conference on computer vision and pattern recognition}, pages 8693--8702, 2024.

\bibitem[Rombach et~al.(2022)Rombach, Blattmann, Lorenz, Esser, and Ommer]{rombach2022high}
Robin Rombach, Andreas Blattmann, Dominik Lorenz, Patrick Esser, and Bj{\"o}rn Ommer.
\newblock High-resolution image synthesis with latent diffusion models.
\newblock In \emph{CVPR}, pages 10684--10695, 2022.

\bibitem[Ruiz et~al.(2023)Ruiz, Li, Jampani, Pritch, Rubinstein, and Aberman]{ruiz2023dreambooth}
Nataniel Ruiz, Yuanzhen Li, Varun Jampani, Yael Pritch, Michael Rubinstein, and Kfir Aberman.
\newblock Dreambooth: Fine tuning text-to-image diffusion models for subject-driven generation.
\newblock In \emph{CVPR}, pages 22500--22510, 2023.

\bibitem[Somepalli et~al.(2024)Somepalli, Gupta, Gupta, Palta, Goldblum, Geiping, Shrivastava, and Goldstein]{somepalli2024measuring}
Gowthami Somepalli, Anubhav Gupta, Kamal Gupta, Shramay Palta, Micah Goldblum, Jonas Geiping, Abhinav Shrivastava, and Tom Goldstein.
\newblock Measuring style similarity in diffusion models.
\newblock \emph{arXiv preprint arXiv:2404.01292}, 2024.

\bibitem[Tan et~al.(2024)Tan, Liu, Yang, Xue, and Wang]{tan2024ominicontrol}
Zhenxiong Tan, Songhua Liu, Xingyi Yang, Qiaochu Xue, and Xinchao Wang.
\newblock Ominicontrol: Minimal and universal control for diffusion transformer.
\newblock \emph{arXiv preprint arXiv:2411.15098}, 3, 2024.

\bibitem[Wang et~al.(2024)Wang, Spinelli, Wang, Bai, Qin, and Chen]{wang2024instantstyle}
Haofan Wang, Matteo Spinelli, Qixun Wang, Xu~Bai, Zekui Qin, and Anthony Chen.
\newblock Instantstyle: Free lunch towards style-preserving in text-to-image generation.
\newblock \emph{arXiv preprint arXiv:2404.02733}, 2024.

\bibitem[Wang et~al.(2025)Wang, Liu, Lin, Liu, Yi, Wang, and Ma]{wang2025omnistyle}
Ye~Wang, Ruiqi Liu, Jiang Lin, Fei Liu, Zili Yi, Yilin Wang, and Rui Ma.
\newblock Omnistyle: Filtering high quality style transfer data at scale.
\newblock In \emph{Proceedings of the Computer Vision and Pattern Recognition Conference}, pages 7847--7856, 2025.

\bibitem[Wei et~al.(2023)Wei, Zhang, Ji, Bai, Zhang, and Zuo]{wei2023elite}
Yuxiang Wei, Yabo Zhang, Zhilong Ji, Jinfeng Bai, Lei Zhang, and Wangmeng Zuo.
\newblock Elite: Encoding visual concepts into textual embeddings for customized text-to-image generation.
\newblock In \emph{CVPR}, pages 15943--15953, 2023.

\bibitem[Wu et~al.(2025{\natexlab{a}})Wu, Li, Zhou, Lin, Gao, Yan, Yin, Bai, Xu, Chen, et~al.]{wu2025qwen}
Chenfei Wu, Jiahao Li, Jingren Zhou, Junyang Lin, Kaiyuan Gao, Kun Yan, Sheng-ming Yin, Shuai Bai, Xiao Xu, Yilei Chen, et~al.
\newblock Qwen-image technical report.
\newblock \emph{arXiv preprint arXiv:2508.02324}, 2025{\natexlab{a}}.

\bibitem[Wu et~al.(2025{\natexlab{b}})Wu, Zheng, Yan, Xiao, Luo, Wang, Li, Jiang, Liu, Zhou, et~al.]{wu2025omnigen2}
Chenyuan Wu, Pengfei Zheng, Ruiran Yan, Shitao Xiao, Xin Luo, Yueze Wang, Wanli Li, Xiyan Jiang, Yexin Liu, Junjie Zhou, et~al.
\newblock Omnigen2: Exploration to advanced multimodal generation.
\newblock \emph{arXiv preprint arXiv:2506.18871}, 2025{\natexlab{b}}.

\bibitem[Wu et~al.(2024)Wu, Ding, Huang, Liu, and He]{wu2024vmix}
Shaojin Wu, Fei Ding, Mengqi Huang, Wei Liu, and Qian He.
\newblock Vmix: Improving text-to-image diffusion model with cross-attention mixing control.
\newblock \emph{arXiv preprint arXiv:2412.20800}, 2024.

\bibitem[Wu et~al.(2025{\natexlab{c}})Wu, Huang, Wu, Cheng, Ding, and He]{wu2025less}
Shaojin Wu, Mengqi Huang, Wenxu Wu, Yufeng Cheng, Fei Ding, and Qian He.
\newblock Less-to-more generalization: Unlocking more controllability by in-context generation.
\newblock \emph{arXiv preprint arXiv:2504.02160}, 2025{\natexlab{c}}.

\bibitem[Wu et~al.(2021)Wu, Nitzan, Shechtman, and Lischinski]{wu2021stylealign}
Zongze Wu, Yotam Nitzan, Eli Shechtman, and Dani Lischinski.
\newblock Stylealign: Analysis and applications of aligned stylegan models.
\newblock \emph{arXiv preprint arXiv:2110.11323}, 2021.

\bibitem[Xiao et~al.(2024)Xiao, Wang, Zhou, Yuan, Xing, Yan, Wang, Huang, and Liu]{xiao2024omnigen}
Shitao Xiao, Yueze Wang, Junjie Zhou, Huaying Yuan, Xingrun Xing, Ruiran Yan, Shuting Wang, Tiejun Huang, and Zheng Liu.
\newblock Omnigen: Unified image generation.
\newblock \emph{arXiv preprint arXiv:2409.11340}, 2024.

\bibitem[Xing et~al.(2024)Xing, Wang, Sun, Wang, Bai, Ai, Huang, and Li]{xing2024csgo}
Peng Xing, Haofan Wang, Yanpeng Sun, Qixun Wang, Xu~Bai, Hao Ai, Renyuan Huang, and Zechao Li.
\newblock Csgo: Content-style composition in text-to-image generation.
\newblock \emph{arXiv preprint arXiv:2408.16766}, 2024.

\bibitem[Xu et~al.(2023)Xu, Liu, Wu, Tong, Li, Ding, Tang, and Dong]{xu2023imagereward}
Jiazheng Xu, Xiao Liu, Yuchen Wu, Yuxuan Tong, Qinkai Li, Ming Ding, Jie Tang, and Yuxiao Dong.
\newblock Imagereward: Learning and evaluating human preferences for text-to-image generation.
\newblock \emph{Advances in Neural Information Processing Systems}, 36:\penalty0 15903--15935, 2023.

\bibitem[Ye et~al.(2023)Ye, Zhang, Liu, Han, and Yang]{ye2023ip}
Hu~Ye, Jun Zhang, Sibo Liu, Xiao Han, and Wei Yang.
\newblock Ip-adapter: Text compatible image prompt adapter for text-to-image diffusion models.
\newblock \emph{arXiv preprint arXiv:2308.06721}, 2023.

\bibitem[Zhai et~al.(2023)Zhai, Mustafa, Kolesnikov, and Beyer]{zhai2023sigmoid}
Xiaohua Zhai, Basil Mustafa, Alexander Kolesnikov, and Lucas Beyer.
\newblock Sigmoid loss for language image pre-training.
\newblock In \emph{Proceedings of the IEEE/CVF international conference on computer vision}, pages 11975--11986, 2023.

\bibitem[Zhang et~al.(2024)Zhang, Song, Liu, Wang, Yu, Tang, Li, Tang, Hu, Pan, et~al.]{zhang2024ssr}
Yuxuan Zhang, Yiren Song, Jiaming Liu, Rui Wang, Jinpeng Yu, Hao Tang, Huaxia Li, Xu~Tang, Yao Hu, Han Pan, et~al.
\newblock Ssr-encoder: Encoding selective subject representation for subject-driven generation.
\newblock In \emph{CVPR}, pages 8069--8078, 2024.

\end{thebibliography}

\clearpage
% \beginappendix
% \clearpage
% \setcounter{page}{12}

% 添加附录标题
\onecolumn
\section*{\centering\LARGE\bfseries USO: Unified Style and Subject-Driven Generation via \\ Disentangled and Reward Learning}
\section*{\centering Appendix}
% 自定义 section 和 subsection 的编号格式
\renewcommand{\thesection}{\Alph{section}}
\renewcommand{\thesubsection}{\Alph{section}.\arabic{subsection}}

\subsection{Experiments Setting}\label{Experiments Setting}
\subsubsection{Implementation Details.}
We begin with FLUX.1 dev~\cite{blackforestlabs_flux} and the SigLIP~\cite{zhai2023sigmoid} pretrained model. For style alignment stage, we train on pairs $\{I^s_{ref}, I_{tgt}\}$ for $23,000$ steps at batch size $16$, learning rate $8e-5$, resolution $768$ and reward steps $S=16,000$. For content-style disentanglement stage, we train on triplets $\{I^c_{ref}, I^s_{ref}, I_{tgt}\}$ for $21,000$ steps at batch size $64$, learning rate $8e-5$, resolution $1024$ and reward steps $S=18,000$. LoRA~\cite{hu2021lora} rank $128$ is used throughout.

\begin{figure*}[h]
\centering
\includegraphics[scale=0.66]{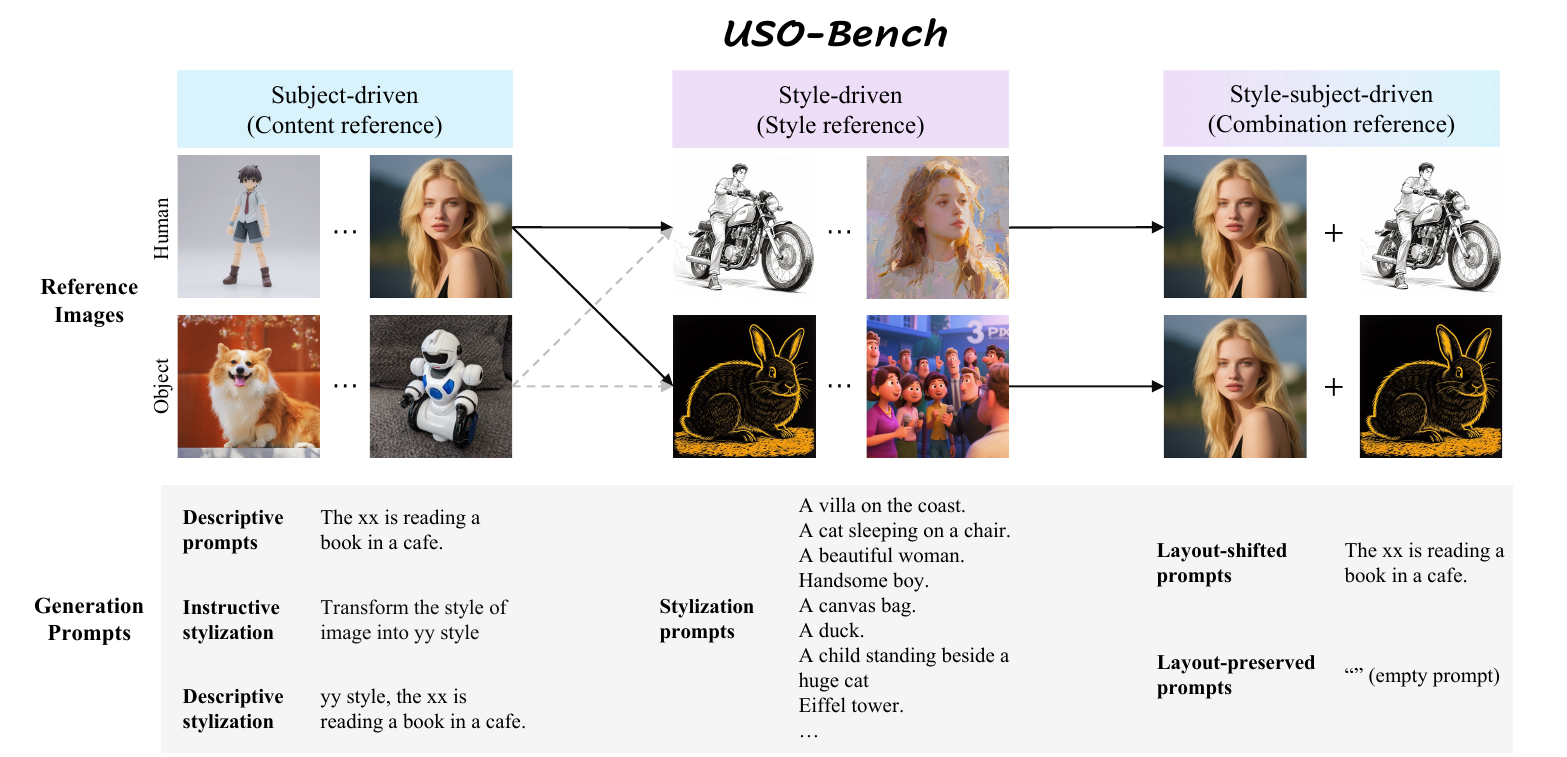}
    \caption{Examples of USO-Bench.}
    \label{sup1:uso-bench}
\end{figure*}

\subsubsection{Details of USO-Bench.}
USO-Bench is built to evaluate subject-driven, style-driven, and joint style-subject-driven generation. As shown in ~\Cref{sup1:uso-bench}, each subject-driven sample uses three prompt types: descriptive, instructive-stylization, and descriptive-stylization. By pairing these prompts with style-reference images from style-driven tasks, we obtain style-subject-driven samples via their Cartesian product. The resulting prompts are further split into layout-shifted and layout-preserved variants.

\begin{table*}[t]
\centering
\small
% \resizebox*{0.425\textwidth}{!}{
% \resizebox*{0.5\textwidth}{!}{
\begin{tabular}{lcccc}
\toprule
\textbf{Method} & \textbf{DINO} $\uparrow$ & \textbf{CLIP-I} $\uparrow$ & \textbf{CLIP-T} $\uparrow$ \\ \toprule
Oracle(reference images) & 0.774 & 0.885 & - \\\hline
Textual Inversion \cite{gal2022image} & 0.569 & 0.780 & 0.255\\
DreamBooth \cite{ruiz2023dreambooth} &0.668  & 0.803 & 0.305 \\
BLIP-Diffusion \cite{li2023blip} & 0.670 & 0.805 & 0.302\\\hline
ELITE \cite{wei2023elite}& 0.647  & 0.772 & 0.296\\
Re-Imagen \cite{chen2022re}& 0.600  & 0.740 & 0.270\\
BootPIG\cite{purushwalkam2024bootpig} & 0.674  & 0.797 & 0.311\\
SSR-Encoder\cite{zhang2024ssr} & 0.612  & 0.821 & 0.308\\
RealCustom++ \cite{huang2024realcustom, mao2024realcustom++}& 0.702  & 0.794 & \textbf{0.318}\\
OmniGen \cite{xiao2024omnigen}& 0.693  & 0.801 & 0.315\\
OminiControl \cite{tan2024ominicontrol}& 0.684  & 0.799 & 0.312\\
FLUX.1 IP-Adapter & 0.582  & 0.820 & 0.288\\
UNO~\cite{wu2025less} & \underline{0.760}  & \underline{0.835} & 0.304\\
\rowcolor{myblue} \textbf{USO (Ours)} & \textbf{0.777} & \textbf{0.838} & \underline{0.317} \\
\bottomrule
\end{tabular}
%}
\caption{Quantitative results for single-subject driven generation on Dreambench~\cite{ruiz2023dreambooth}.}
\label{tab1:sup_dreambench}
\end{table*}

\subsection{More Results}\label{Experiments Setting}
\subsubsection{Quantitative Evaluation on DreamBench~\cite{ruiz2023dreambooth}.}
To further assess USO, we evaluate it on DreamBench~\cite{ruiz2023dreambooth} in addition to USO-Bench. Following UNO~\cite{wu2025less}, we generate six images per prompt, yielding 4,500 image groups across all subjects. As shown in ~\Cref{tab1:sup_dreambench}, USO achieves the highest CLIP-I and DINO scores, and with a CLIP-T score of 0.317, it trails the top result (0.318) by only a narrow margin. These results demonstrate USO’s superior subject consistency among state-of-the-art methods.
\begin{table*}[t]
    \centering
    \renewcommand{\arraystretch}{0.9} % 调整行间距
     \resizebox{0.9\linewidth}{!}{
    \begin{tabular}{lc}
        \toprule
        \textbf{Scenarios} & \textbf{Prompt} \\
        \midrule
        Subject/Identity Driven Generation & (1) "The girl is riding a bike in the street." \\
                & (2) "The man is driving a car in the street." \\
                & (3) "A sophisticated gentleman exuding confidence. He is dressed in a 1990s brown plaid jacket with a \\
                & high collar, paired with a dark grey turtleneck. His trousers are tailored and charcoal in color, \\
                & complemented by a sleek leather belt. The background showcases an elegant library with \\
                & bookshelves, a marble fireplace, and warm lighting, creating a refined and cozy atmosphere. \\
                & His relaxed posture and casual hand-in-pocket stance add to his composed and stylish demeanor" \\ 
                & (4) "The woman is reading a book in a cafe." \\
        \midrule
        Subject/Identity Driven Stylization & (1) "Sketch style, a bowl with a mountain in the background." \\
              & (2) "Illustration style, a dog on the beach." \\
              & (3) "Transform to Picasso's style of work, Cubism." \\
              & (4) "Ghibli style, The woman rides a deer in the forest." \\
        \midrule
        Style Driven Generation & (1)"A shark." \\
              & (2) "Small boat in the lake." \\    
              & (3) "A beautiful woman." \\    
              & (4) "The top chef is stir-frying in the kitchen." \\    
        \midrule
        Multi-style Driven Generation & (1) "A beautiful woman." \\
              & (2) "A duck, with words read "USO",  "inspires creativity"." \\    
              & (3) "A man." \\    
        \midrule
        Style-subject Driven Generation & (1) "" \\
           (Layout-preserved)   & (2) "" \\
           & (3) "" \\
        \midrule
        Style-subject Driven Generation & (1) "A toy in the jungle." \\
           (Layout-shifted)   & (2) "A cat on the beach." \\
           & (3) "The woman gave an impassioned speech on the podium." \\
        \bottomrule
    \end{tabular}
    }
    \caption{Text prompts used in \Cref{teaser}.}\label{tab:teaser}
\end{table*}  

\begin{figure*}[h]
\centering
\includegraphics[width=0.9\textwidth]{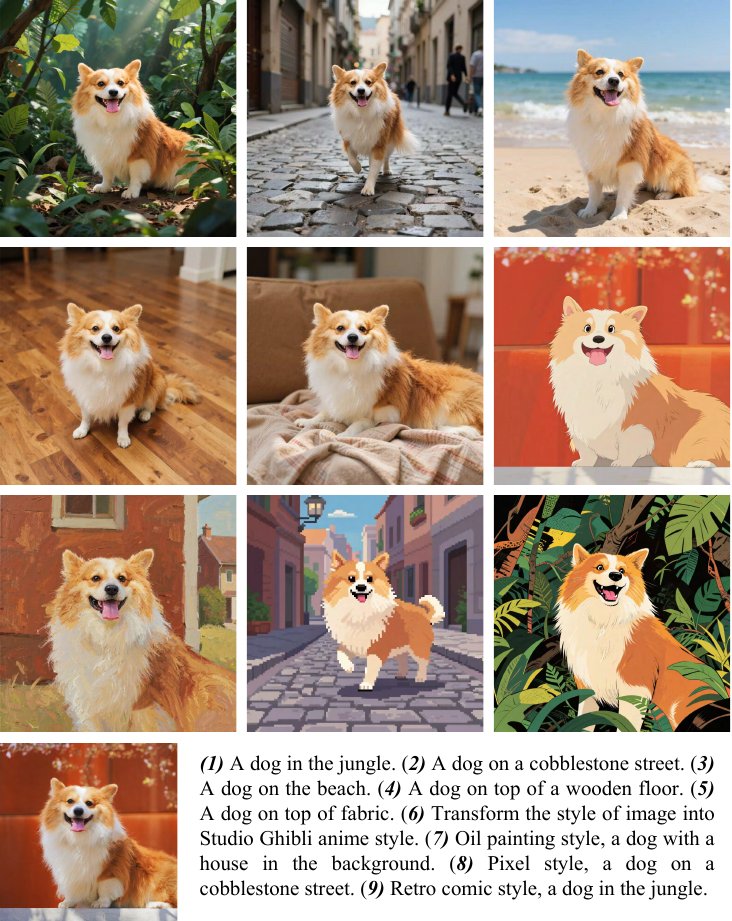}
    \caption{More results on subject-driven generation.}
    \label{sup3:subject1}
\end{figure*}

\begin{figure*}[h]
\centering
\includegraphics[width=0.9\textwidth]{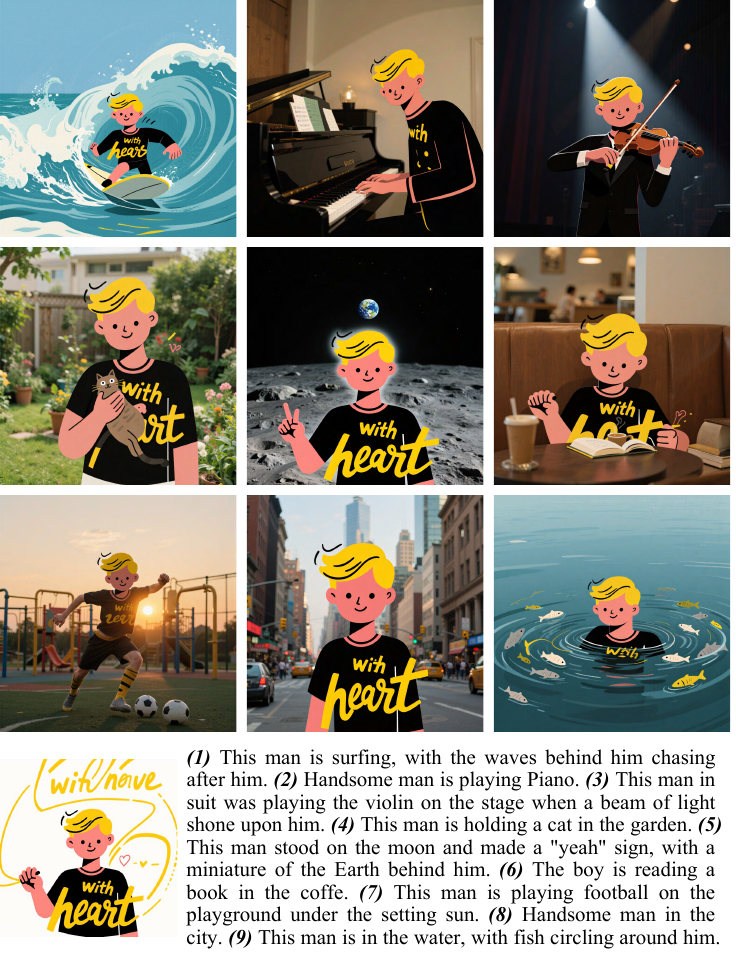}
    \caption{More results on subject-driven generation.}
    \label{sup3:subject2}
\end{figure*}

\begin{figure*}[h]
\centering
\includegraphics[width=0.9\textwidth]{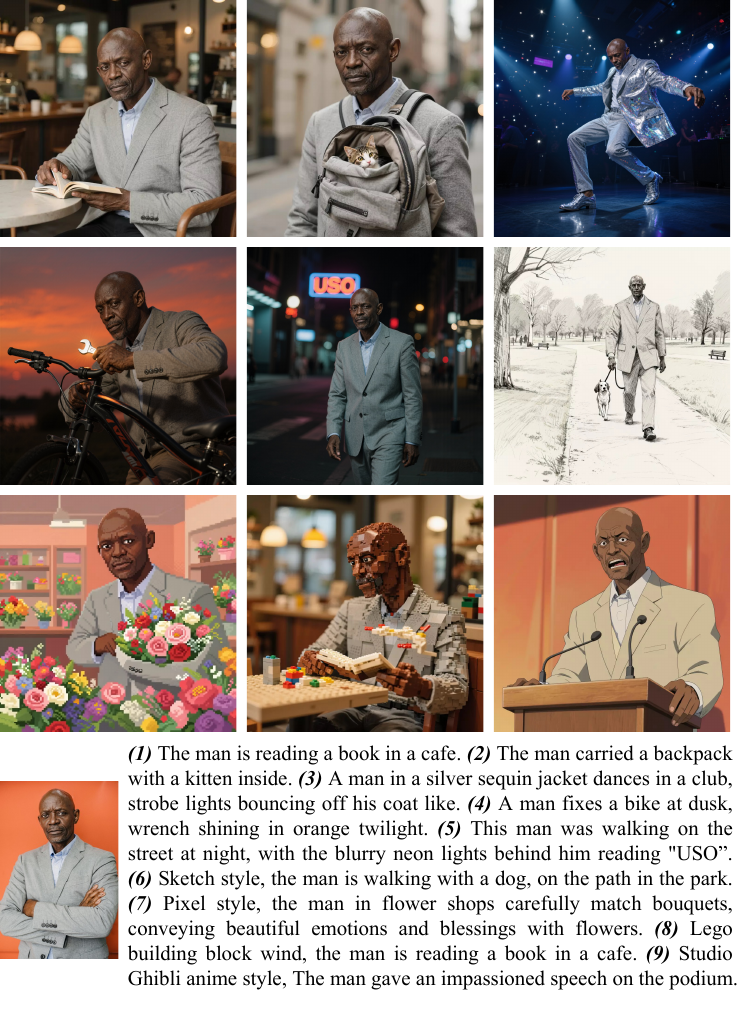}
    \caption{More results on identity-driven generation.}
    \label{sup3:subject3}
\end{figure*}

\begin{figure*}[h]
\centering
\includegraphics[width=0.9\textwidth]{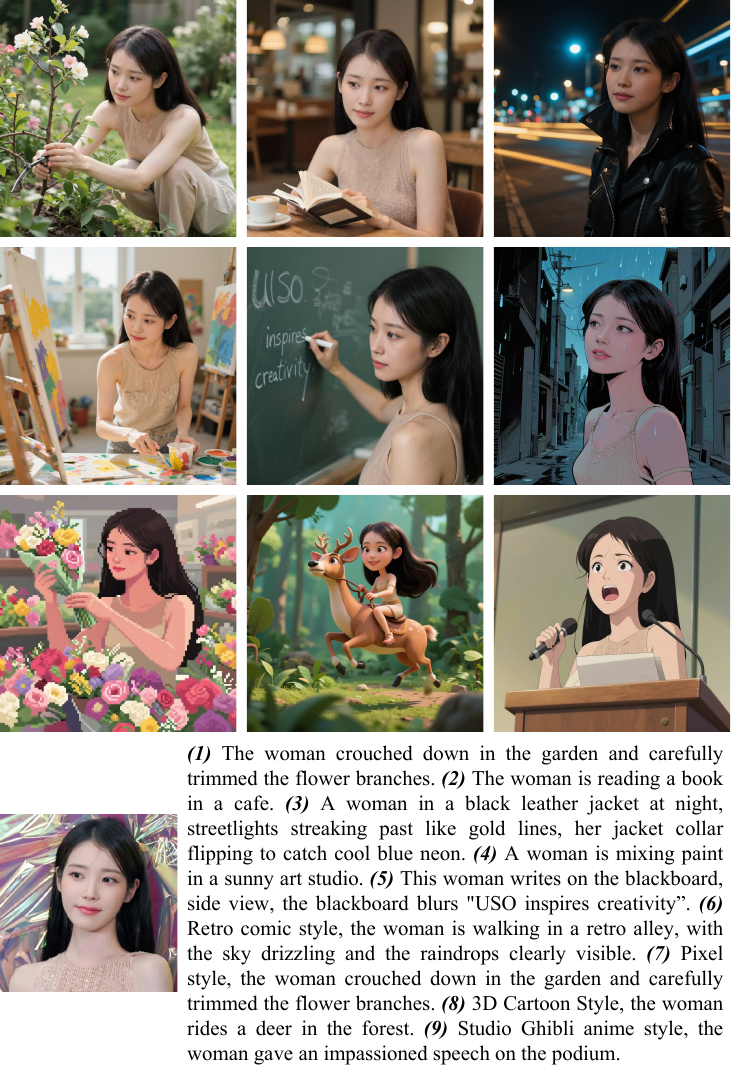}
    \caption{More results on identity-driven generation.}
    \label{sup3:subject4}
\end{figure*}

\begin{figure*}[h]
\vspace*{-0.11\textwidth}
\hspace*{-0.055\textwidth}
\includegraphics[width=1.1\textwidth]{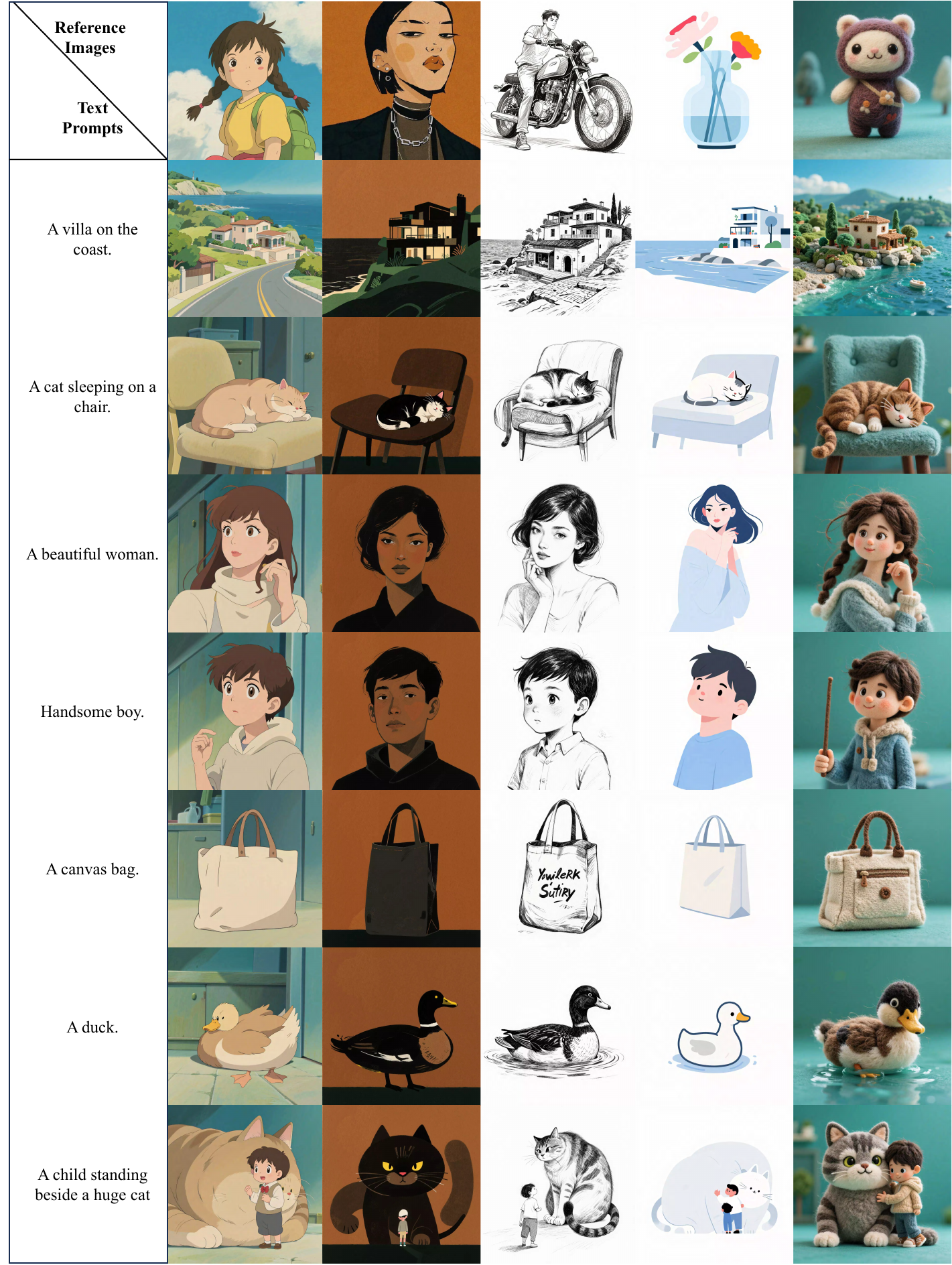}
    \caption{More results on style-driven generation.}
    \label{sup4:style1}
\end{figure*}

\begin{figure*}[h]
\vspace*{-0.11\textwidth}
\hspace*{-0.055\textwidth}
\includegraphics[width=1.1\textwidth]{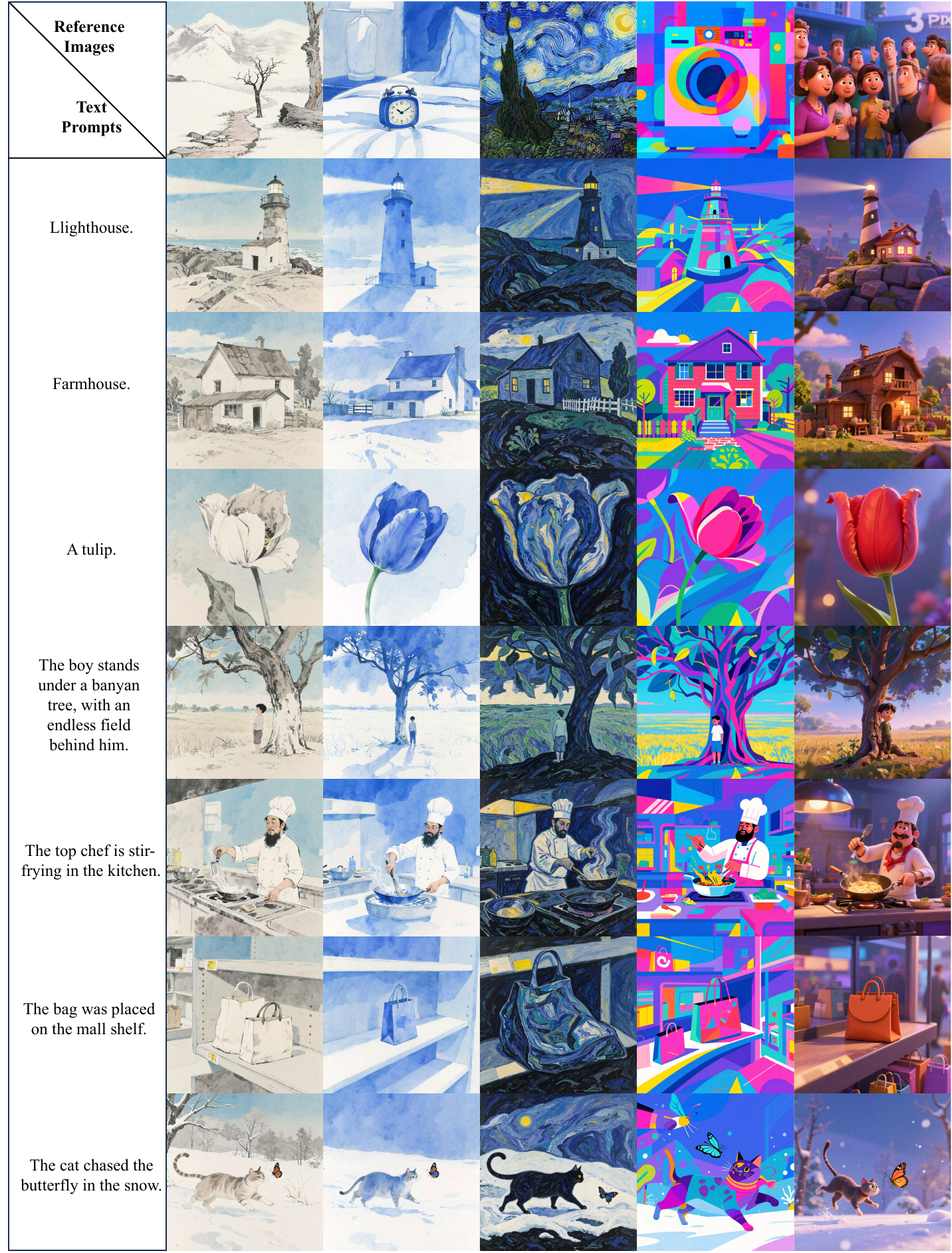}
    \caption{More results on style-driven generation.}
    \label{sup4:style2}
\end{figure*}

\begin{figure*}[h]
\vspace*{-0.11\textwidth}
\hspace*{-0.055\textwidth}
\includegraphics[width=1.05\textwidth]{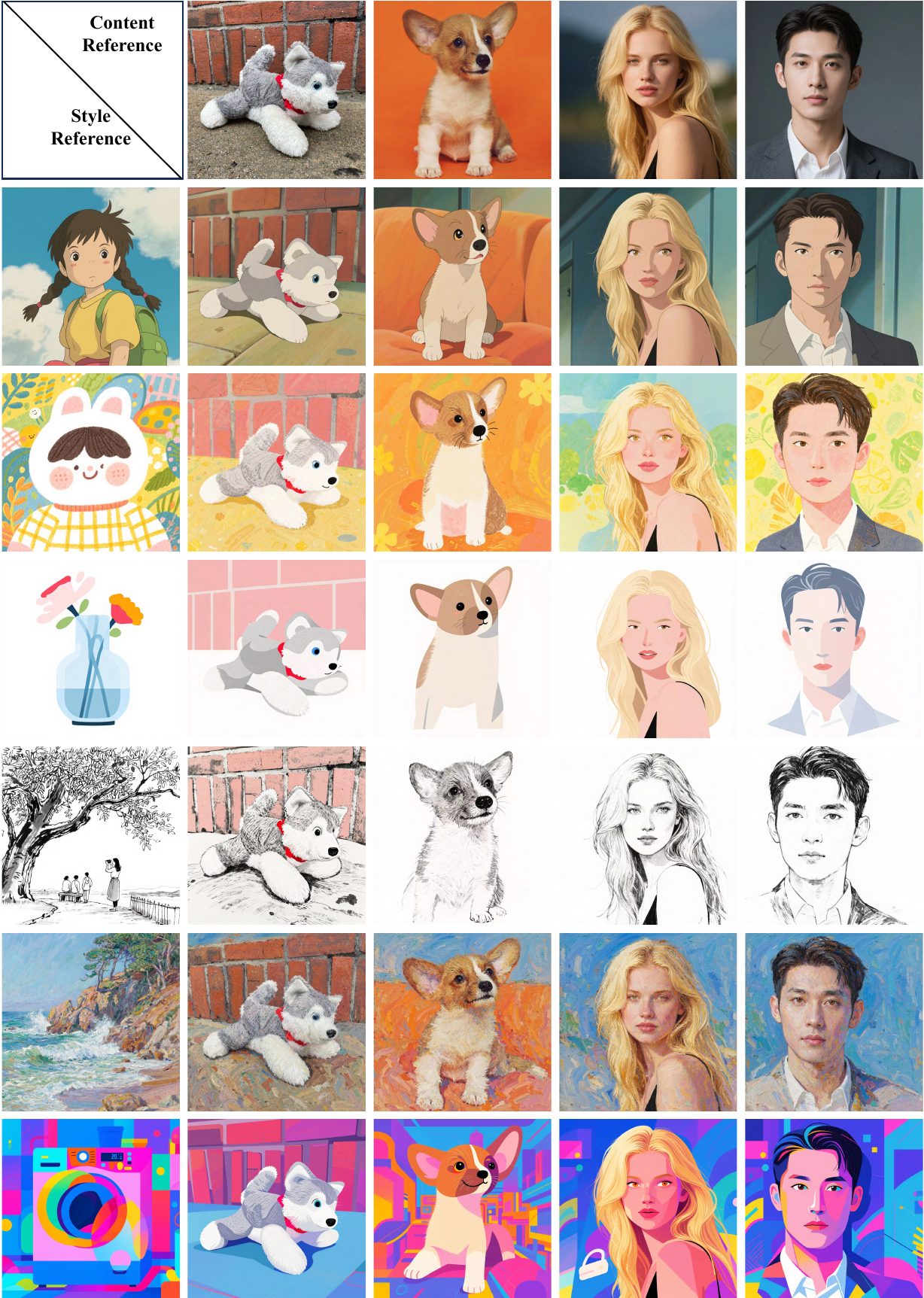}
    \caption{More results on style-subject-driven generation. We set prompt to empty for layout-preserved generation.}
    \label{sup4:ipstyle1}
\end{figure*}

\begin{figure*}[h]
\vspace*{-0.11\textwidth}
\hspace*{-0.055\textwidth}
\includegraphics[width=1.05\textwidth]{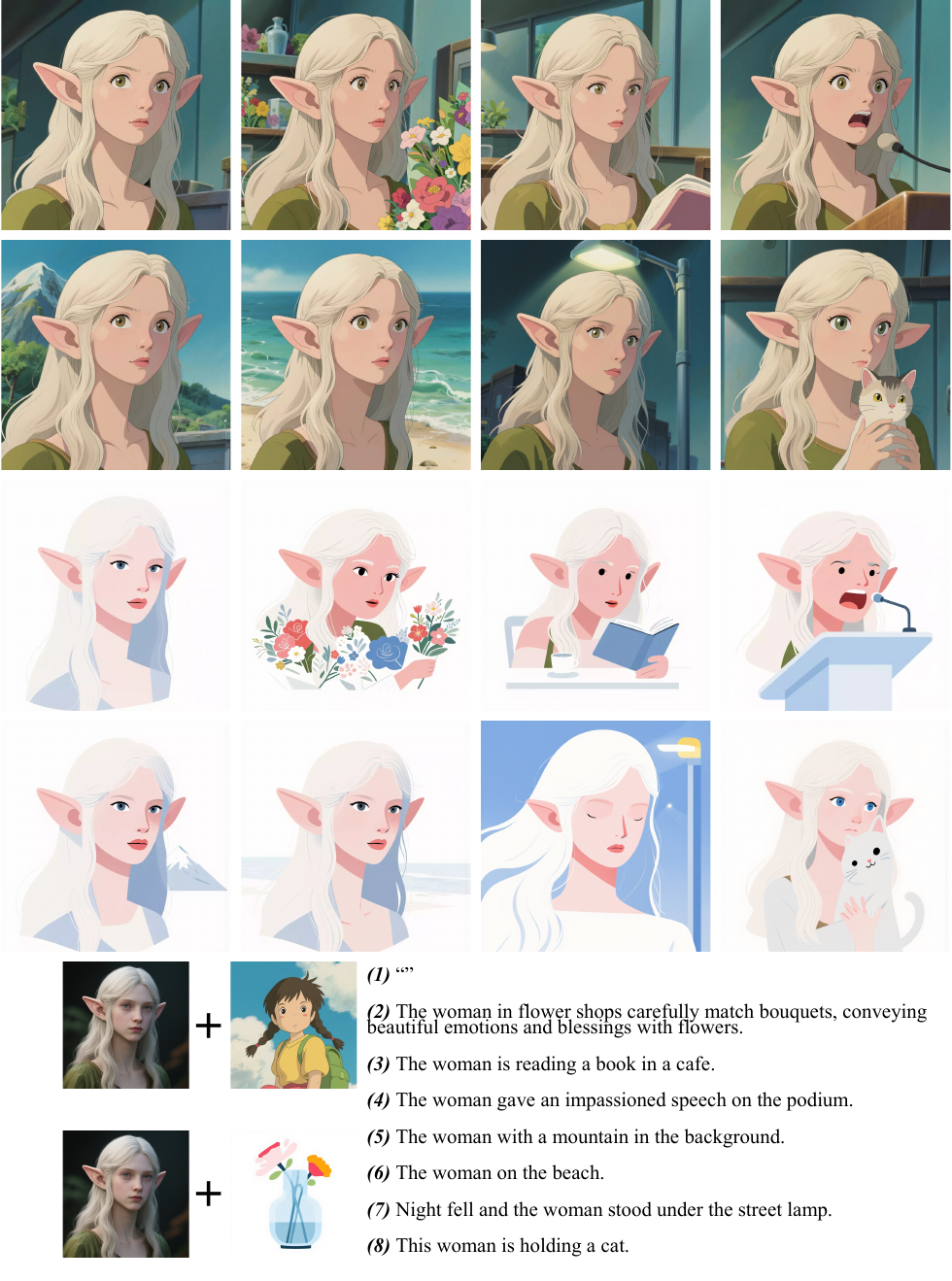}
    \caption{More results on style-subject-driven generation. USO supports any subject combined with any style in any scenario.}
    \label{sup4:ipstyle2}
\end{figure*}

\subsubsection{Additional Results.}
We present additional qualitative results from USO:
\begin{itemize}
\item From \Cref{sup3:subject1,sup3:subject2,sup3:subject3,sup3:subject4}, USO demonstrates the ability to extract task-relevant content features while maintaining subject consistency across diverse textual prompts—capabilities that prior work typically treats as isolated tasks (e.g., subject-driven generation, instruction-based stylized editing, and identity preservation).
\item In \Cref{sup4:style1,sup4:style2}, USO exhibits high stylistic fidelity, capturing both fine-grained characteristics (e.g., brushwork and material textures) and abstract artistic styles—far beyond simple color transfer.
\item In \Cref{sup4:ipstyle1,sup4:ipstyle2}, USO freely combines arbitrary subjects with arbitrary styles, supporting both layout-preserving and layout-shifting generations.
\end{itemize}
\end{document}